\documentclass[final,1p,times,numbers,sort&compress]{elsarticle}

\usepackage{xcolor}
\usepackage{graphicx}
\usepackage{mathbbold}
\usepackage{subfigure}
\usepackage{amsmath}
\usepackage{amssymb}
\usepackage{chemsym}
\usepackage{slashbox}
\usepackage{bm}
\usepackage{multirow}
\usepackage{algorithm}
\usepackage{algorithmic}

\newtheorem{theorem}{Theorem}[section]

\newproof{pf}{Proof}

\journal{}

\begin{document}

\begin{frontmatter}



\title{Image denoising based on improved data-driven sparse representation}


\author[adr1]{Dai-Qiang Chen \corref{cor1}}

\cortext[cor1]{Corresponding author}
\ead{chener050@sina.com}


\address[adr1]{Department of Mathematics, College of Biomedical Engineering, Third Military Medical University, Chongqing 400038,
Chongqing, People's Republic of China}

\begin{abstract}

Sparse representation of images under certain transform domain has been playing a fundamental role in image restoration tasks. One such representative method is the widely used wavelet tight frame. Instead of adopting fixed filters to sparsely model any input image, a data-driven tight frame was proposed and shown to be very efficient for image denoising very recently. However, in this method the number of framelet filters used for constructing a tight frame is the same as the length of filters. In fact, through further investigation it is found that part of these filters are unnecessary and even harmful to the recovery effect due to the influence of noise. Therefore, an improved data-driven sparse representation systems constructed with much less number of filters are proposed. Numerical results on denoising experiments demonstrate that the proposed algorithm overall outperforms the original tight frame construction scheme on both the recovery quality and computational time.
\end{abstract}


\end{frontmatter}


\section{Introduction}\label{sec1}

Image denoising, which aims to estimate the clean image from its noise-corrupted observation, is a fundamental topic in the field of computer vision and image processing. Two types of commonly encountered noise are additive white Gaussian noise (AWGN) and multiplicative noise (MN). AWGN is often introduced in the optical imaging due to the thermal motion of electron in camera sensors or circuits, and MN (also called speckle) often appears in coherent imaging systems which are different from the usual optical imaging, e.g., synthetic aperture radar (SAR) or ultrasound imaging. By taking the logarithmic transform of the data, the MN or speckle noise model can be converted into a tractable additive noise \cite{JMIV:MN,GRSL:MN,JSC:MN}. Then many denoising techniques which are widely used for AWGN can also be applied for multiplicative noise removal \cite{JMIV:MN,GRSL:MN,TIP:MN}. In this article, we focus our attention on AWGN denoising problem.

In the last several years, the sparsity-inducing prior based on transformed frames has been playing a very
important role in the development of effective image denoising methods. The fundamental idea behind is that the interest image or image patches can be sparsely represented by properly designed transformed frames. One such popular method is the wavelet tight frames \cite{ACHA:Framelets,Notes:Wavelet}, which use fixed functions such as the linear or cubic B-spline as the generators and can be obtained by the unitary extension principle (UEP). Wavelet tight frames are widely used for image restoration tasks due to its strong ability of sparsely approximating piecewise smooth regions and preserving the jump discontinuities, and the low computational cost \cite{ACHA:frameletinpainting,SIIMS:LBFrame,JAMS:TVWave,TIP:Blind,SIIMS:GILAM}. However, the structure of images varies greatly in practice, and a tight frame working well for one type of images may be not suitable for another. In order to properly represent the textures and tiny details including in the image, many approaches of learning an over-complete dictionary from the image \cite{NC:DLA,Book:Elad,TIP:KSVD,MMS:MKSVD} were developed in recent years. One of the representative work along this direction is the famous K-SVD method \cite{TIP:KSVD}, which devises an alternating minimization algorithm to learn an over-complete dictionary from the noisy image patches, and use the learned dictionary to sparsely model and denoise images. These dictionary learning methods improve the quality of denoised images. However, the corresponding computational cost is too high due to the fact that the sparse coding of a large number of image patches is required during each iteration. Recently, the patch-dictionary learning methods were combined with another widely used nonlocal techniques \cite{MMS:NonLocal,SP:nlpro}, and various novel image restoration models \cite{CVPR:CSR,PR:TSPCA,ITP:NCSR,ICCV:NLSM,TIP:BM3d,TIP:LSSC,CVPR:WNNM} were generated. These nonlocal sparsity representation methods were among the current state-of-the-art approaches in the quality of recovered images. However, the corresponding computational cost is really too high due to the computation of patch-similarity weights, patches group and sparse coding for a large number of image patches. In fact, only the computation of similarity weights will spend much time.

Very recently, an adaptive tight frame construction scheme was proposed to overcome the drawback of the traditional wavelet tight frames \cite{ACHA:DDTF}. Differently from the existing tight framelet methods, the main idea of the new approach is to learn a tight frame from the input image. Through minimizing a problem with respect to the framelet filters and the canonical frame coefficient vectors, a data-driven tight frame can be efficiently learned by several iteration steps of simple singular value decomposition (SVD) and hard thresholding operator. Numerical experiments demonstrate that, with comparable performance in image denoising, the proposed data-driven tight frame construction scheme runs much faster than the general dictionary learning methods such as the classical K-SVD method. The proposed variational model in \cite{ACHA:DDTF} is a non-convex model and the convergence of the corresponding alternating minimization algorithm is further investigated in the recent literature \cite{ACHA:ConverageDDTF}, which proves the sub-sequence convergence property of the algorithm. The data-adaptive tight framelet has also been directly applied to remove the multiplicative noise by simply using the logarithmic transform of the data \cite{GRSL:MN}.

In the above-mentioned data-driven tight frame construction algorithm, the framelet filters are updated by the simple SVD of a matrix with the size of $p^{2}\times p^{2}$ at each iteration, where $p\times p$ is the size of the chosen filters. Therefore, the number of framelet filters learned from the input image is the same as the length of the support set of filters, which is also equal to $p^{2}$. However, in the implementation process of the proposed algorithm we observe that a majority of these learned framelet filters are unnecessary and even harmful to the denoised results due to the fact that the noise obviously influences the learned filters that correspond to the features that seldom appear in the input image. Based on this observed phenomena, we divide the matrix used for generating the framelet filters into two part, which correspond to the signal subspace and noise subspace respectively. Then we further propose an improved data-driven sparse representation method with much less number of filters learned from the bases of the signal subspace. Due to the removal of filters which are useless for the sparse representation of the image features, the computational cost is further reduced and the quality of denoising results is not damaged at the same time. The proposed data-driven framelet construction scheme is compared with the original algorithm with application to AWGN denoising problem.

The rest of this paper is organized as follows. In section \ref{sec2} we briefly review the classical K-SVD algorithm and the recently proposed data-driven tight frame construction algorithm. In section \ref{sec3}, in order to overcome the drawback of the previous work, we propose a new filters learning method, which excludes the inessential framelet filters. The convergence of the proposed algorithm is further investigated. In section \ref{sec4} the numerical examples on AWGN denoising are reported to compare the proposed algorithm with the original data-driven tight frame construction scheme.

\section{Data-driven tight frame construction scheme}\label{sec2}

In this section, we briefly introduce some preliminaries of wavelet tight frames, and then review and compare the classical K-SVD algorithm and the recently proposed data-driven tight frame construction scheme.

Starting with a finite set of generators $\Psi=\{\Psi_{1}, \cdots, \Psi_{m}\}$, one wavelet frame for $L_{2}(\mathbb{R})$ is just the affine system generated by the shifts and dilations of $\Psi$, i.e., it is defined by
\[
X(\Psi)=\left\{\Psi_{q,j,k}, ~~1\leq q\leq m, j\in \mathbb{Z}, k\in \mathbb{Z}\right\}
\]
where $\Psi_{q,j,k}=2^{j/2}\Psi_{q}(2^{j}\cdot-k)$.
Such system $X(\Psi)$ is called a tight frame of $L_{2}(\mathbb{R})$ if it satisfies
$f=\sum_{\psi\in \Psi}\langle f, \psi\rangle \psi$ for any $f\in L_{2}(\mathbb{R})$. A tight frame has two associated operators: one is the analysis operator $W$ defined by
\[
W: u\in L_{2}(\mathbb{R})\rightarrow \left\{\langle u, \Psi_{q,j,k}\rangle\right\},
\]
and the other is the adjoint operator $W^{T}$ (also called the synthesis operator) defined by
\[
W^{T}: \left\{\alpha_{q,j,k}\right\} \rightarrow \sum \alpha_{q,j,k} \Psi_{q,j,k}\in L_{2}(\mathbb{R}).
\]
The construction of wavelet tight frame is using the multi-resolution analysis (MRA). In the discrete setting, it can be generated by the shifts and dilations of a set of masks or filters $\{h_{1}, \cdots,  h_{m}\}$ associated with the corresponding generators $\Psi$. For more details refer to \cite{ACHA:Framelets,Notes:Wavelet}. The framelets for $L_{2}(\mathbb{R}^{2})$ can be easily constructed by taking tensor products of $1$-dimensional framelets.

However, in the tight frame systems introduced above, the filters are fixed and unable to be adaptively adjusted according to the input image. Therefore, the approaches of learning an over-complete dictionary from the image were widely developed in recent years. One of the representative work is the K-SVD method \cite{TIP:KSVD}. Here we give a brief review on this method.

Let $f$ and $g$ denote the noise-free image and the noisy image respectively, and $\left\{\overrightarrow{f_{i}}\right\}_{i=1}^{N}\in \mathbb{R}^{p^{2}}$  be the set of image patches of size $p\times p$ densely extracted from the image $f$. Let $\tilde{D}=\left(\overrightarrow{d_{1}}, \overrightarrow{d_{2}}, \cdots, \overrightarrow{d_{m}}\right)\in \mathbb{R}^{p^{2}\times m}$ denote the dictionary whose column vectors represent the dictionary atoms. Then the K-SVD method for image denoising is expressed by the following minimization problem :

\begin{equation}\label{equ2.1rev1}
\begin{split}
\min_{f,v,\tilde{D}}\frac{1}{2}\|g-f\|_{2}^{2} + \sum_{i}\lambda_{i}\|v_{i}\|_{0} + \mu \sum_{i}\|\tilde{D}v_{i}-\overrightarrow{f_{i}}\|_{2}^{2}.
\end{split}
\end{equation}
where $v$ is the the sparse coding coefficient composed by $v_{i}$, and $v_{i}$ denotes the expansion coefficient vector of the $i$-th image patch over the dictionary $\tilde{D}$.

The performance of the K-SVD method outperform the wavelet tight frames in image denoising due to the fact that the repeating textures or features of images are likely to be captured by the learned dictionary atoms. However, the corresponding minimization problem (\ref{equ2.1rev1}) is difficult to be solved efficiently. An alternating iterative algorithm is adopted to update the values of $f, v$ and $\tilde{D}$ respectively in literature \cite{TIP:KSVD}. Specifically, the sparse coding coefficient is calculated in each iteration by the orthogonal matching pursuit (OMP), which is quite slow and accounts for most of the computational amount of this method.

Very recently, Cai et.al \cite{ACHA:DDTF} further proposed a variational model to learn a tight frame system from the image itself. The tight frames considered there are single-level un-decimal discrete wavelet systems generated by the data-driven filters $\{h_{1}, \cdots,  h_{m}\}$. For each $h_{i}$, let $\overrightarrow{h_{i}}$ be its column vector formed by concatenating all its columns. Define $H=\left[\overrightarrow{h_{1}}, \overrightarrow{h_{2}}, \cdots, \overrightarrow{h_{m}}\right]$, and $W(H)$ and $W^{T}(H)$ be the analysis operator and the synthesis operator generated by the filters $H$ respectively. The problem of learning the filters $H$ from the image is still difficult for the arbitrary $m$. Therefore, the authors in \cite{ACHA:DDTF} consider a special case of $m=p^{2}$ and each $h_{i}$ be a real-valued filter with support on $\mathbb{N}^{2}\cap [1,p]^{2}$.
In this situation, it has been proved that the constraint $W^{T}(H)W(H)=I$ exists as long as $H^{T}H=\frac{1}{p^{2}}I$. The interested reader can refer to Proposition $3$ of \cite{ACHA:DDTF} for details.

Based on this conclusion, a novel variational model for data-driven tight frame construction is proposed in \cite{ACHA:DDTF}, which can be briefly summarized as follows:

\begin{equation}\label{equ2.3}
\min\limits_{v, \{h_{i}\}_{i=1}^{m}} \|v-W(H)g\|^{2}_{2} + \lambda_{0}^{2} \|v\|_{0}, ~~\textrm{s.t.}~~H^{T}H=\frac{1}{p^2}I
\end{equation}
where $v$ is the coefficient vector that sparsely approximates the tight frame coefficient $W(H) g$.
Compared with the minimization problem (\ref{equ2.1rev1}), it is observed that the coefficient $v$ in problem (\ref{equ2.3}) can be simply obtained by the hard thresholding operator, and hence the new method avoids the huge computational burden of updating the coefficient $v$ in the K-SVD method. Intuitively, the K-SVD method  (\ref{equ2.1rev1}) can be seen as a synthesis-based model for sparse representation; and the new variational model (\ref{equ2.3}) is just an analysis-based model which can be solved more easily and thus the computational efficiency can be improved greatly.

The problem (\ref{equ2.3}) can be rewritten in the matrix form. For each patch $\overrightarrow{g_{i}}$, let $\overrightarrow{v_{i}}$ denote the vector corresponding to $H^{T}\overrightarrow{g_{i}}\in \mathbb{R}^{p^{2}}$. Define

\begin{equation}\label{equ2.4}
\left\{\begin{array}{lll}G=\frac{1}{p}\left(\overrightarrow{g_{1}}, \overrightarrow{g_{2}}, \cdots, \overrightarrow{g_{N}}\right)\in \mathbb{R}^{p^{2}\times N}, &~ &~ ~\\
V=\left(\overrightarrow{v_{1}}, \overrightarrow{v_{2}}, \cdots, \overrightarrow{v_{N}}\right)\in \mathbb{R}^{p^{2}\times N},
&~ &~ ~\\ D=p\left(\overrightarrow{h_{1}}, \overrightarrow{h_{2}}, \cdots, \overrightarrow{h_{p^{2}}}\right)\in \mathbb{R}^{p^{2}\times p^{2}}.& ~
&~
\end{array} \right.
\end{equation}
Then problem (\ref{equ2.3}) is equal to
\begin{equation}\label{equ2.5}
\min\limits_{V, D} \|V-D^{T}G\|^{2}_{F} + \lambda^{2} \|V\|_{0}, ~~\textrm{s.t.}~~D^{T}D=I.
\end{equation}
where $\lambda$ represents some fixed regularization parameter. In literature \cite{ACHA:DDTF}, this problem is solved via an alternating minimization scheme between the coefficients $V$ and the learning filters $D$. More specifically, given the current iteration value $D_{k}$ of the variable $D$, it is easy to deduce that the optimal value $V$ of the minimization problem (\ref{equ2.5}) can be simply obtained by the hard thresholding operator. In what follows, fixed the value of $V$ as $V_{k+1}$, the next iteration update for $D$ is given by the minimization problem

\begin{equation}\label{equ2.7}
\min\limits_{D} \|V_{k+1}-D^{T}G\|^{2}_{F}, ~~\textrm{s.t.}~~D^{T}D=I.
\end{equation}
Let $U_{k}\Sigma_{k}X_{k}^{T}$ be the SVD of $G V_{k+1}^{T}\in \mathbb{R}^{p^{2}\times p^{2}}$ such that $G V_{k+1}^{T}=U_{k}\Sigma_{k}X_{k}^{T}$. It has been shown in \cite{ACHA:DDTF} that the minimization problem (\ref{equ2.7}) has a closed solution denoted by

\begin{equation}\label{equ2.8}
D_{k+1}=U_{k}X_{k}^{T}.
\end{equation}
It is observed that the filters set $D$ associated with the framelet can be obtained by a simple SVD of a matrix of small size $p^{2}\times p^{2}$.

\section{Improved data-driven sparse representation method}\label{sec3}

\subsection{Motivation and model description}\label{subsec3.1}

In the above-mentioned algorithm, the matrix $D$ associated with the framelet filters is updated by the iterative formula (\ref{equ2.8}), where $U_{k}$ and $X_{k}$ are given by the SVD of $G V_{k+1}^{T}$. Assume that $\Sigma_{k}=diag\{r_{i}\}$, where $\{r_{i}\}_{i=1}^{p^{2}}$ denotes the singular value of $G V_{k+1}^{T}$ in a descending order. Let $U_{k}=\left\{u_{k}^{1}, u_{k}^{2}, \cdots, u_{k}^{p^{2}}\right\}$ and $X_{k}=\left\{x_{k}^{1}, x_{k}^{2}, \cdots, x_{k}^{p^{2}}\right\}$. Then $G V_{k+1}^{T}$ can be reformulated as
\begin{equation}\label{equ3.1}
G V_{k+1}^{T} = \sum_{i=1}^{p^{2}}r_{i}u_{k}^{i}(x_{k}^{i})^{T} = \underbrace{\sum_{i=1}^{s}r_{i}u_{k}^{i}(x_{k}^{i})^{T}}_{signal~subspace} + \underbrace{\sum_{i=s+1}^{p^{2}}r_{i}u_{k}^{i}(x_{k}^{i})^{T}}_{noise~subspace}.
\end{equation}
In the denoising problem, the input image $g$ used for generating the matrix $G$ is contaminated by noise, thus so is the matrix $G V_{k+1}^{T}$.
Therefore, we can divide the whole space spanned by the bases $\left\{u_{k}^{i}(x_{k}^{i})^{T}\right\}_{i=1}^{p^{2}}$ into two orthogonal subspace: the signal subspace which is spanned by the bases corresponding to the $s (s<p^{2})$ largest singular values; and the noise subspace which is spanned by the bases corresponding to the rest singular values. Then in the formula (\ref{equ3.1}) the matrix $G V_{k+1}^{T}$ is divided into two part--one is the signal component obtained by projection on the signal subspace, the other is the noise component obtained by projection on the noise subspace.

Rewrite $D^{k+1}=\sum \limits_{i=1}^{s}u_{k}^{i}(x_{k}^{i})^{T} + \sum \limits_{i=s+1}^{p^{2}}u_{k}^{i}(x_{k}^{i})^{T}$. The last part of $\sum \limits_{i=s+1}^{p^{2}}u_{k}^{i}(x_{k}^{i})^{T}$ can be regarded as the filters component mainly learned from the noise component of the input image, and hence maybe useless and even harmful to the final denoised results. Therefore, it should be removed in the filters learning process.

\begin{figure}
  \centering
  \subfigure[]{
    \label{fig3.1:subfig:a}
    \includegraphics[width=1.5in,clip]{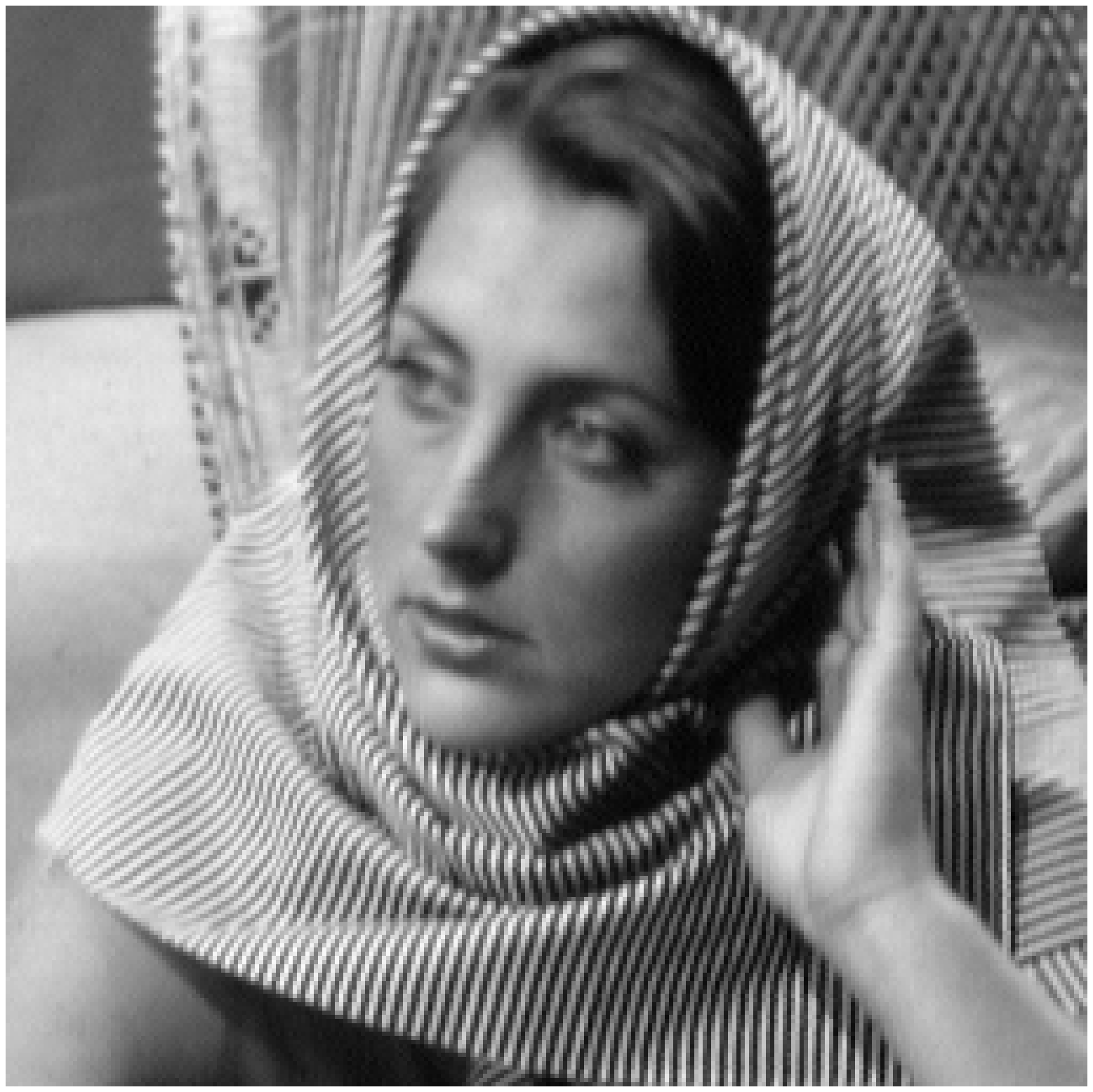}}
  \subfigure[]{
    \label{fig3.1:subfig:b}
    \includegraphics[width=1.5in,clip]{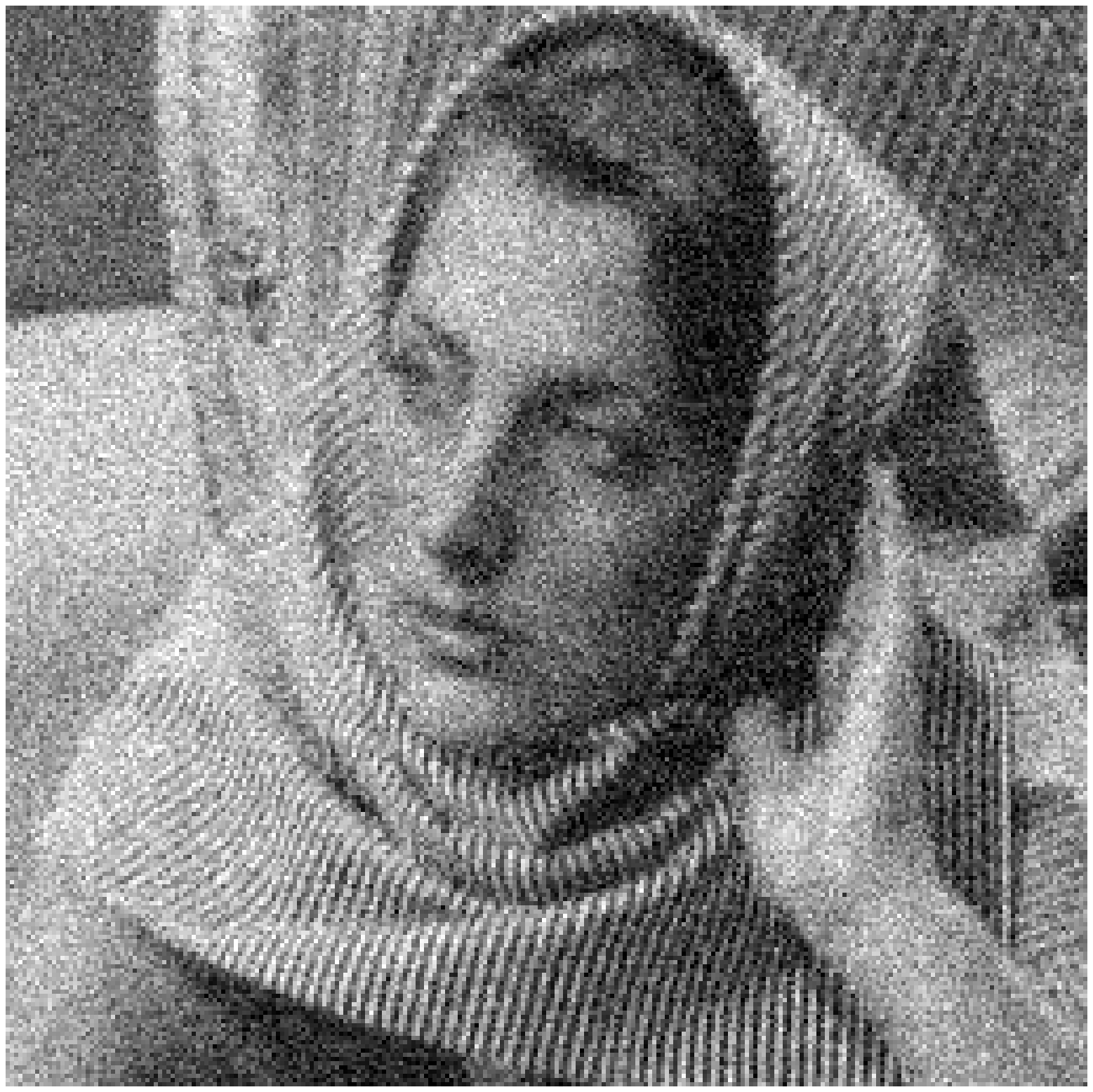}}
  \subfigure[]{
    \label{fig3.1:subfig:c}
    \includegraphics[width=1.8in,clip]{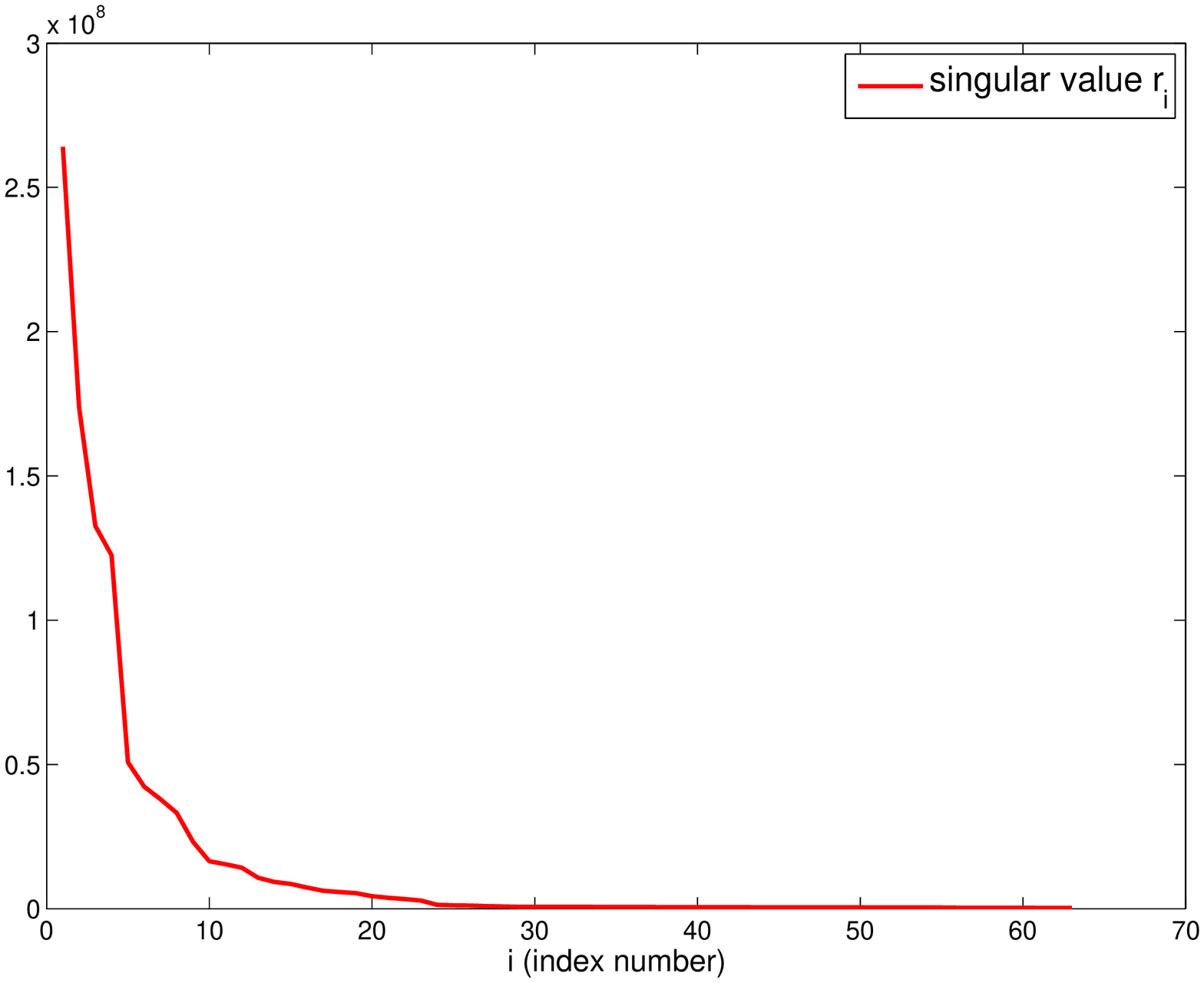}} \\
  \subfigure[]{
    \label{fig3.1:subfig:d}
    \includegraphics[width=1.5in,clip]{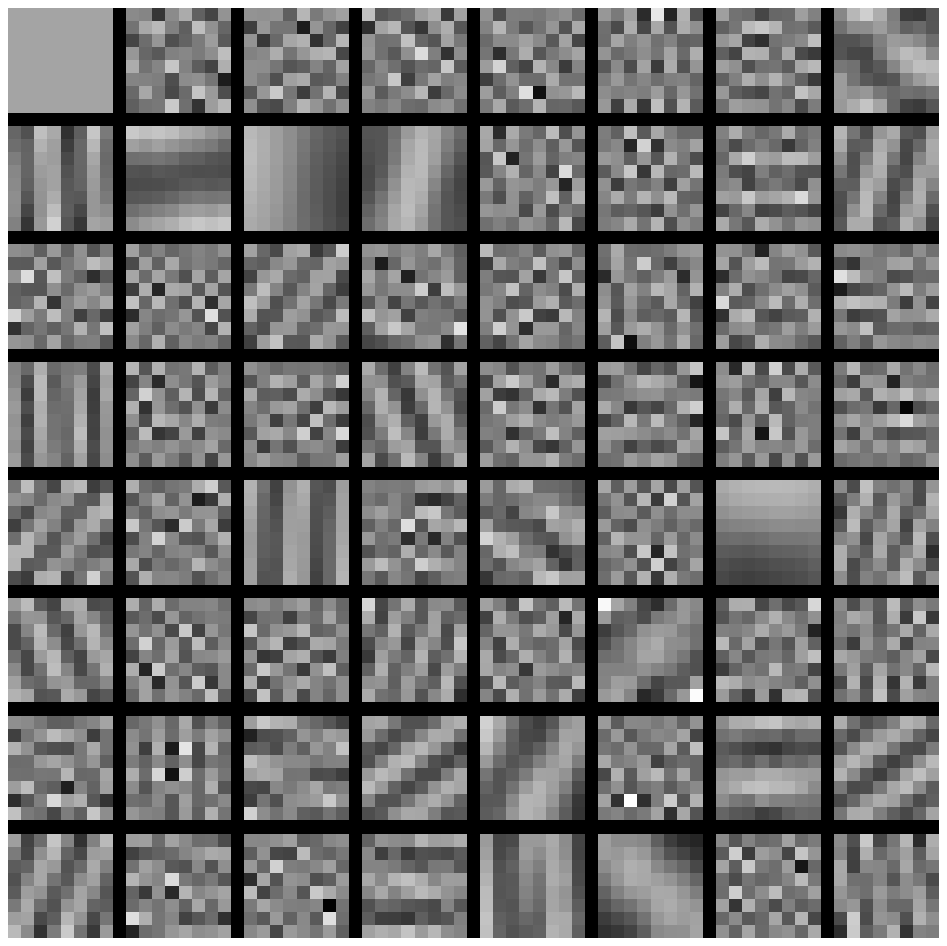}} =
  \subfigure[]{
    \label{fig3.1:subfig:e}
    \includegraphics[width=1.5in,clip]{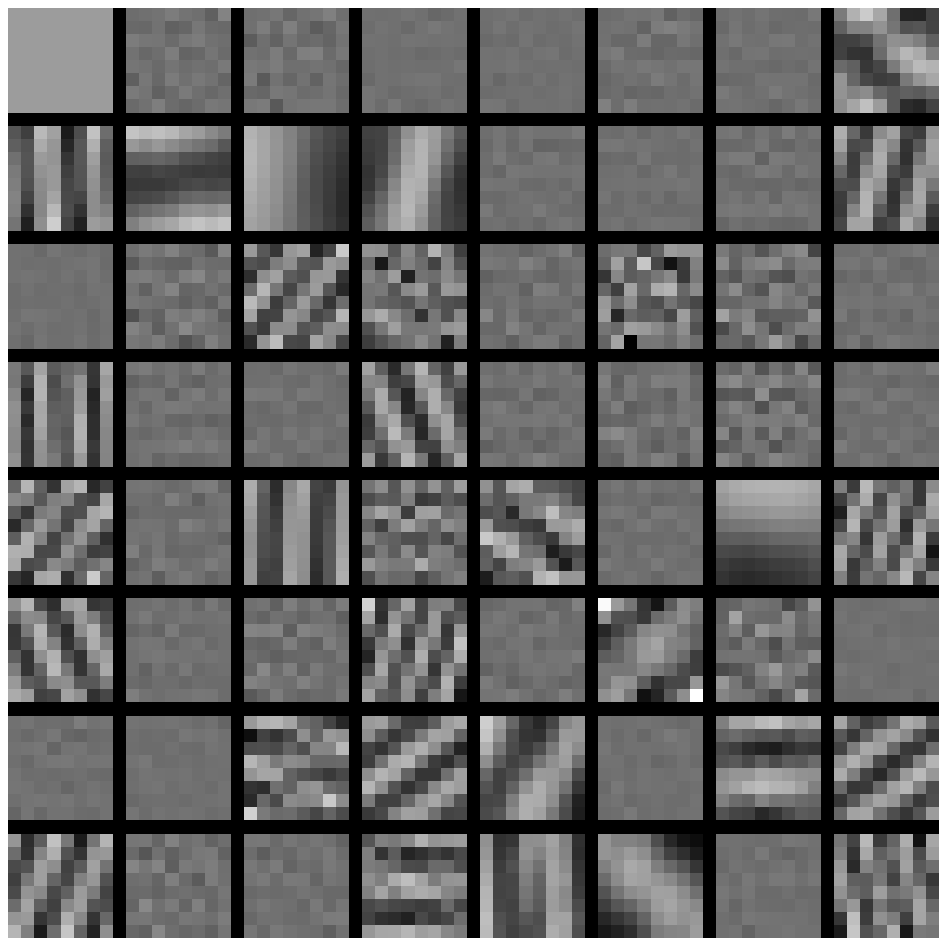}} +
  \subfigure[]{
    \label{fig3.1:subfig:f}
    \includegraphics[width=1.5in,clip]{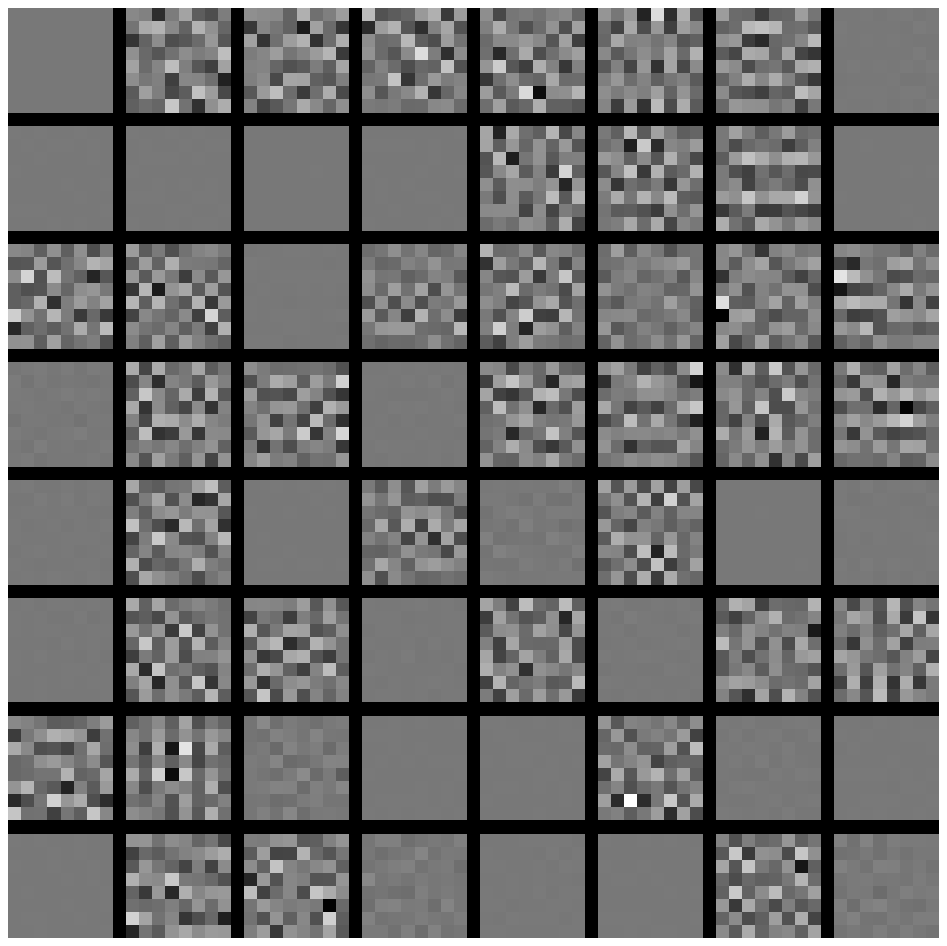}}
\caption{(a) The original Barbara image, (b) the noisy image, (c) the final singular values of $G V_{k+1}^{T}$ after $25$ iteration steps, (d) corresponding adaptive tight frame filters, (e) frame filters generated by the bases of signal subspace, (f) frame filters generated by the bases of noise subspace.
}
\label{fig3.1}
\end{figure}

To illustrate this conclusion more clearly, we use the Barbara image of size $256\times 256$ contaminated by zero-mean AWGN of $\sigma=25$ as an example, where $\sigma$ denotes the standard deviation of the noise. Figure \ref{fig3.1:subfig:c} shows the curve of the singular values of $G V_{K}^{T} (K=25)$. It is observed that the singular values $r_{i}$ decrease very fast with the increase of the index $i$. This phenomenon implies that the framelet filters corresponding to the last several singular values of $G V_{k+1}^{T}$ will be influenced by the noise more obviously compared with those corresponding to the larger singular values.
{In fact, the observation in Figure \ref{fig3.1:subfig:c} reveals the low-rank properties of the features of images. It is well known that the low-rank property has been widely utilized for developing image restoration models \cite{TIP:LSSC,CVPR:WNNM,SIAMImage:Lowrank}. Numerical examples in section \ref{sec4} also verify the low-rank properties.}
Figure \ref{fig3.1:subfig:d} presents the obtained adaptive tight frame filters generated by $\sum_{i=1}^{p^{2}}u_{k}^{i}(x_{k}^{i})^{T}$. It is found that many filters included in the figure are likely to be contaminated by the noise, e.g., the second to seventh filters in the first row, and the fifth to seventh filters in the second row of Figure \ref{fig3.1:subfig:d}. This is due to the fact that these filters are generated by the bases corresponding to the last several singular values of $G V_{k+1}^{T}$ and hence influenced by the noise obviously. Figures \ref{fig3.1:subfig:e} and \ref{fig3.1:subfig:f} also show the frame filters generated by $\sum_{i=1}^{s}u_{k}^{i}(x_{k}^{i})^{T}$ and $\sum_{i=s+1}^{p^{2}}u_{k}^{i}(x_{k}^{i})^{T}$, where we choose $s=30$ (The selection of $s$ is influenced by the amount and pattern of repeating textures or features and the noisy level. Through many trials for different values of $s$, we found that $s=30$ is a suitable value for separating the signal subspace and noise subspace of the Barbara image). It is clearly observed that the filters contained in the noisy subspace have been severely influenced by the noise, and hence are not suitable for the sparse representation of the image.

Motivated by the above analysis, we propose a new strategy to remove the useless framelet filters and develop an improved data-driven sparse representation method. Assume that only $s (s<p^{2})$ filters including in the filters group $H$ are suitable for the sparse representation of the noisy image. Without loss of generality, we group filters $H$ into two classes: $H_{1}=\left[\overrightarrow{h_{1}}, \overrightarrow{h_{2}}, \cdots, \overrightarrow{h_{s}}\right]\in \mathbb{R}^{p^{2}\times s}$ and $H_{2}=\left[\overrightarrow{h_{s+1}}, \overrightarrow{h_{s+2}}, \cdots, \overrightarrow{h_{p^{2}}}\right]\in \mathbb{R}^{p^{2}\times (p^{2}-s)}$, which correspond to the bases of the signal subspace and the noise subspace respectively,
and hence only $H_{1}$ is suitable for the sparse modeling of the noisy image. Then similarly to (\ref{equ2.3}), a variational model for the sparse representation of the image is proposed as follows

\begin{equation}\label{equ3.3rev}
\begin{split}
\min\limits_{v, H_{1}} \|v-W(H_{1})g\|^{2}_{2} + \lambda_{0}^{2} \|v\|_{0}, \\ ~~\textrm{s.t.}~~H_{1}^{T}H_{1}=\frac{1}{p^2}I.
\end{split}
\end{equation}
The proposed model (\ref{equ3.3rev}) can only demonstrate that the image can be sparsely represented by the filters $H_{1}$. However, it cannot guarantee that $H_{1}$ is generated by the bases of the signal subspace. Due to the fact that $H_{1}$ captures the strong features of the image, the following constraint
\begin{equation}\label{equ3.1rev}
\max\limits_{H_{1}}\|W(H_{1})g\|^{2}_{2}
\end{equation}
should be satisfied for the filters $H_{1}$. It is noticed that
\[
\begin{split}
\|W(H_{1})g\|^{2}_{2}+\|W(H_{2})g\|^{2}_{2}=\\
g^{T}W(H_{1})^{T}W(H_{1})g+g^{T}W(H_{2})^{T}W(H_{2})g\\
=g^{T}W(H)^{T}W(H)g =g^{T}g=\textrm{const}.
\end{split}
\]
Therefore, $\max\limits_{H_{1}}\|W(H_{1})g\|^{2}_{2}$ is equal to
\begin{equation}\label{equ3.2rev}
\min\limits_{H_{2}}\|W(H_{2})g\|^{2}_{2}.
\end{equation}
The model (\ref{equ3.2rev}) demonstrates that $H_{2}$ corresponds to the projection onto the noise subspace, in other words, $H_{1}$ corresponds to the projection onto the signal subspace.

Combining the models (\ref{equ3.3rev}) and (\ref{equ3.2rev}) we further propose a variational model for adaptive filters learning as follows

\begin{equation}\label{equ3.2}
\begin{split}
\min\limits_{v, H_{1}, H_{2}} \|v-W(H_{1})g\|^{2}_{2} + \|W(H_{2})g\|^{2}_{2} + \lambda_{0}^{2} \|v\|_{0}, \\ ~~\textrm{s.t.}~~H^{T}H=\frac{1}{p^2}I.
\end{split}
\end{equation}
{It is noteworthy that the second term in (\ref{equ3.2}) plays the role of eliminating the filters from the noise subspace, and hence cannot be removed. More precisely, in the example shown in Figure \ref{fig3.1}, if we choose $H1=\left[\overrightarrow{h_{1}}, \overrightarrow{h_{2}}, \cdots, \overrightarrow{h_{30}}\right]\in \mathbb{R}^{p^{2}\times 30}$, and $H2=\left[\overrightarrow{h_{1}}, \overrightarrow{h_{2}}, \cdots, \overrightarrow{h_{15}}, \overrightarrow{h_{49}}, \overrightarrow{h_{50}}, \cdots, \overrightarrow{h_{64}}\right]\in \mathbb{R}^{p^{2}\times 30}$, through direct computation we can observe that
\[
\frac{\left\{\|v_{H1}-W(H1)g\|^{2}_{2} + \lambda_{0}^{2} \|v_{H1}\|_{0}\right\}}{\left\{\|v_{H2}-W(H2)g\|^{2}_{2} + \lambda_{0}^{2} \|v_{H2}\|_{0}\right\}}<1
\]
where $v_{H1}$ and $v_{H2}$ represent the coefficients corresponding to $H1$ and $H2$ respectively.
The above result implies $H2$ that contains the last 15 filters contaminated by noise is superior to $H1$ in the problem
\[
\min_{v_1, H_1} ||v_1 - W(H_1) g||_2^2 + \lambda_{0}^{2} \|v\|_{0} ~s.t.~ H_1^{T}H_1=\frac{1}{p^2}I, ~H_1\in \mathbb{R}^{p^{2}\times 30}.
\]
This is due to that $\|\overrightarrow{h_{i}}^T G\|_2<\|\overrightarrow{h_{j}}^T G\|_2$ for $h_{i}$ and $h_{j}$ that belong to the noise subspace and signal subspace respectively, and hence $||\textbf{0} - \overrightarrow{h_{i}}^T G||_2^2$ may be smaller than  $||v_{j} - \overrightarrow{h_{j}}^T G||_2^2+\lambda^2 ||v_j||_0$. Therefore, the extra term $\|W(H_{2})g\|^{2}_{2}$ is necessary.}

In what follows, we further reformulate the minimization problem (\ref{equ3.2}) in the matrix form. Define

\begin{equation}\label{equ3.3}
\left\{\begin{array}{lll}G=\frac{1}{p}\left(\overrightarrow{g_{1}}, \overrightarrow{g_{2}}, \cdots, \overrightarrow{g_{L}}\right)\in \mathbb{R}^{p^{2}\times N}, &~ &~ ~\\
V=\left(\overrightarrow{v_{1}}, \overrightarrow{v_{2}}, \cdots, \overrightarrow{v_{L}}\right)\in \mathbb{R}^{s\times N},
&~ &~ ~\\ D_{1}=p\left(\overrightarrow{h_{1}}, \overrightarrow{h_{2}}, \cdots, \overrightarrow{h_{s}}\right)\in \mathbb{R}^{p^{2}\times s},
&~ &~ ~\\ D_{2}=p\left(\overrightarrow{h_{s+1}}, \overrightarrow{h_{s+2}}, \cdots, \overrightarrow{h_{p^{2}}}\right)\in \mathbb{R}^{p^{2}\times (p^{2}-s)},
&~ &~ ~\\ D=(D_{1}, D_{2})\in \mathbb{R}^{p^{2}\times p^{2}}.
\end{array} \right.
\end{equation}
Then problem (\ref{equ3.2}) can be rewritten as

\begin{equation}\label{equ3.4}
\min\limits_{D, V} \|V-D_{1}^{T}G\|^{2}_{F} +  \|D_{2}^{T}G\|^{2}_{F} + \lambda^{2} \|V\|_{0}, ~~\textrm{s.t.}~~D^{T}D=I.
\end{equation}
The established model can be solved by the alternating minimization scheme between $V$ and $D$. Specifically,
given the current estimate $(V_{k}, D_{k})$, the next iteration updates its value via the following scheme:

\begin{equation}\label{equ3.5}
\left\{\begin{array}{lll}V_{k+1}=\textrm{arg}\min \limits_{V\in \mathbb{R}^{s\times N}}\|V-(D_{1})_{k}^{T}G\|^{2}_{F} + \lambda^{2} \|V\|_{0},
&~ &~ ~\\ D_{k+1}=\textrm{arg}\min \limits_{D\in \mathbb{R}^{p^{2}\times p^{2}}} \|V_{k+1}-D_{1}^{T}G\|^{2}_{F} +  \|D_{2}^{T}G\|^{2}_{F}, \\ ~~\textrm{s.t.}~~D^{T}D=I.
\end{array} \right.
\end{equation}
The first sub-problem of (\ref{equ3.5}) has a closed solution as follows:

\begin{equation}\label{equ3.6}
V_{k+1}=\mathcal{H}_{\lambda}((D_{1})_{k}^{T}G)
\end{equation}
where $\mathcal{H}_{\lambda}: \mathbb{R}^{s\times N}\rightarrow \mathbb{R}^{s\times N}$ is the hard thresholding operator defined by

\begin{equation}\label{equ3.7}
[\mathcal{H}_{\lambda}(U)]_{i,j}=\begin{cases}U_{i,j},\quad   \textrm{if} ~|U_{i,j}|> \lambda,\\
  0, \quad \ \ \textrm{otherwise}.
\end{cases}
\end{equation}
The objective function of the second sub-problem of (\ref{equ3.5}) can be reformulated as

\begin{equation}\label{equ3.8}
\begin{aligned}
\|V_{k+1}-D_{1}^{T}G\|^{2}_{F} +  \|D_{2}^{T}G\|^{2}_{F}\\
= \textrm{Tr}((V_{k+1}-D_{1}^{T}G)^{T}(V_{k+1}-D_{1}^{T}G)) + \textrm{Tr}(G^{T}D_{2}D_{2}^{T}G)\\
= \textrm{Tr}(V_{k+1}^{T}V_{k+1}) + \textrm{Tr}(G^{T}DD^{T}G) - 2\textrm{Tr}(D_{1}^{T}GV_{k+1}^{T}) \\
= \textrm{Tr}(V_{k+1}^{T}V_{k+1}) + \textrm{Tr}(G^{T}G) - 2\textrm{Tr}(D_{1}^{T}GV_{k+1}^{T}).
\end{aligned}
\end{equation}
Note that the third equality of (\ref{equ3.8}) is from the fact that $DD^{T}=I$, which is derived directly from $D^{T}D=I$. Therefore, the optimal value $(D_{1})_{k+1}$ is the solution of the following problem

\begin{equation}\label{equ3.9}
\max_{D_{1}} \textrm{Tr}(D_{1}^{T}GV_{k+1}^{T}) ~~\textrm{s.t.}~~D_{1}^{T}D_{1}=I.
\end{equation}
Suppose that the SVD of $G V_{k+1}^{T}\in \mathbb{R}^{p^{2}\times s}$ is $\tilde{U}_{k+1}\tilde{\Sigma}_{k+1}\tilde{X}_{k+1}^{T}$, where $\tilde{U}_{k+1}\in \mathbb{R}^{p^{2}\times s}$, $\tilde{\Sigma}_{k+1}\in \mathbb{R}^{s\times s}$, and $\tilde{X}_{k+1}\in \mathbb{R}^{s\times s}$. Then based on Theorem 4 in \cite{ACHA:DDTF} we know that the minimization problem (\ref{equ3.9}) has a closed solution

\begin{equation}\label{equ3.10}
(D_{1})_{k+1} = \tilde{U}_{k+1}\tilde{X}_{k+1}^{T}.
\end{equation}
The value of $(D_{2})_{k+1}$ should satisfy that $D_{k+1}^{T}D_{k+1}=I$. In other words, it should be the orthocomplement of $(D_{1})_{k+1}$. Since it represents the useless filters and is independent with the update of $V$, the corresponding computation can be neglected here.
Following the above analysis, we obtain the improved data-driven sparse representation method, which is summarized in Algorithm 1. After the learned filters set $(D_{1})_{K}$ is obtained, the denoised image can be given by

\begin{equation}\label{equ3.11}
g_{K} = (W((D_{1})_{k}))^{T}(\mathcal{H}_{\tilde{\lambda}}(W((D_{1})_{k})g)).
\end{equation}
where $\tilde{\lambda}$ is the threshold determined by the noise level.

\begin{algorithm}[htb]
\caption{ Construction of the data-driven filters for sparse representation}
\begin{algorithmic}[1]
\REQUIRE
the noisy $g$.\\
\textbf{Output:} data-driven filters set $D_{1}$ defined by $p\left(\overrightarrow{h_{1}}, \overrightarrow{h_{2}}, \cdots, \overrightarrow{h_{s}}\right)\in \mathbb{R}^{p^{2}\times s}$. \\
\textbf{Initialization}: set the number of filters $s$, and initial filter matrix $D_{0}$; construct the patch matrix $G$ defined by (\ref{equ3.3}). \\
\textbf{Iteration}: For $k=0, 1, \cdots, K-1$ \\
~~~(\romannumeral1) update $V$: \\
~~~~~~$V_{k+1}=\mathcal{H}_{\lambda}((D_{1})_{k}^{T}G)$; \\
~~~(\romannumeral2) update $D_{1}$: \\
~~~~~~compute the SVD decomposition of $G V_{k+1}^{T} = \tilde{U}_{k+1}\tilde{\Sigma}_{k+1}\tilde{X}_{k+1}^{T}$; and update $D_{1}$ as $(D_{1})_{k+1} = \tilde{U}_{k+1}\tilde{X}_{k+1}^{T}$. \\
\end{algorithmic}
\end{algorithm}

\subsection{Further analysis of the proposed model}\label{subsec3.2}

In this section, we further analyze the convergence of the proposed algorithm. In recent years, the convergence of the alternating proximal minimization algorithm for non-convex minimization problems has been widely investigated \cite{MOR:KLProperty,MP:KLProperty}. One of the widely used tool is the Kurdyka-Lojasiewicz (K-L) property, see \cite{ACHA:ConverageDDTF,MOR:KLProperty,SIAMOptim:KLProperty} for more details.
In literature \cite{ACHA:ConverageDDTF}, the authors used the results presented in \cite{MOR:KLProperty} to further investigate the convergence of the data-driven tight frame construction scheme introduced in section \ref{sec2}. The convergence of the proposed algorithm can be proved very similarly.

Denote the convex set
\[
\begin{split}
\Omega_{\tilde{V}}=\{v_{i,j}~|~ v_{i,j}=0 ~\textrm{for any}~ i>s\}, \\
\Omega_{D}=\{D~|~ D^{T}D=I\}
\end{split}
\]
for any $\tilde{V}\in \mathbb{R}^{p^{2}\times N}$. Then the proposed model (\ref{equ3.4}) can be reformulated as

\begin{equation}\label{equ3.12}
\min\limits_{D, V\in \Omega_{\tilde{V}}} \|\tilde{V}-D^{T}G\|^{2}_{F} + \lambda^{2} \|\tilde{V}\|_{0}, ~~\textrm{s.t.}~~D^{T}D=I.
\end{equation}
Define
\begin{equation}\label{equ3.13}
\begin{split}
f(\tilde{V})=\lambda^{2} \|\tilde{V}\|_{0}+\iota_{\Omega_{\tilde{V}}}(\tilde{V}), ~Q(\tilde{V}, D) \\
=\|\tilde{V}-D^{T}G\|^{2}_{F}, ~h(D)=\iota_{\Omega_{D}}(D)
\end{split}
\end{equation}
where $\iota_{\Omega}$ denotes the indicative function of the set $\Omega$, i.e., $\iota_{\Omega}(A)=0$ if $A\in \Omega$, and $+\infty$ otherwise. Then, the minimization problem (\ref{equ3.12}) can be rewritten as

\begin{equation}\label{equ3.14}
\min\limits_{D, \tilde{V}} \mathcal{L}(\tilde{V}, D)=f(\tilde{V})+h(D)+Q(\tilde{V}, D).
\end{equation}
Similarly to the proof in literature \cite{ACHA:ConverageDDTF}, we can easily derive the following convergence result. Due to the proof requires only minor changes to that presented in \cite{ACHA:ConverageDDTF}, we omit it here due to limited space.

\begin{theorem}\label{the1}
(The convergence of Algorithm $1$) The sequence $\{D^{k}, V^{k}\}$ generated by Algorithm $1$ has at least one limit point. Let $(D^{\ast}, V^{\ast})$ be any limit point of the sequence $\{D^{k}, V^{k}\}$. Denote $\tilde{V}^{\ast}=(V^{\ast};\textbf{0})\in \mathbb{R}^{p^{2}\times N}$. Then $(D^{\ast}, \tilde{V}^{\ast})$ is a stationary point of (\ref{equ3.14}).
\end{theorem}

However, the above conclusion only illustrates the sub-sequence convergence property of the proposed algorithm. The sequence convergence property can be guaranteed by coupling this alternating minimization method with a proximal term. See \cite{ACHA:ConverageDDTF,MOR:KLProperty,MP:KLProperty} for more details. In this setting, the proposed iteration scheme can be modified as

\begin{equation}\label{equ3.15}
\left\{\begin{array}{lll}\tilde{V}_{k+1}=\textrm{arg}\min \limits_{\tilde{V}}\mathcal{L}(\tilde{V}, D^{k})+\lambda_{k}\|\tilde{V}-\tilde{V}_{k}\|_{F}^{2},
&~ &~ ~\\ D_{k+1}=\textrm{arg}\min \limits_{D} \mathcal{L}(\tilde{V}_{k+1}, D)+\mu_{k}\|D-D_{k}\|_{F}^{2}
\end{array} \right.
\end{equation}
where $\lambda_{k}, \mu_{k}\in(c_{1}, c_{2})$ and $c_{1}, c_{2}$ are two positive constants.
Consequently, the iteration scheme of Algorithm $1$ can be modified as

\begin{equation}\label{equ3.16}
\left\{\begin{array}{lll}V_{k+1}=\mathcal{H}_{\lambda/\sqrt{1+\lambda_{k}}}\left(\frac{(D_{1})_{k}^{T}G + \lambda_{k} V_{k}}{1+\lambda_{k}}\right),
&~ &~ ~\\ (D_{1})_{k+1}=\tilde{U}_{k+1}\tilde{X}_{k+1}^{T}
\end{array} \right.
\end{equation}
where $\tilde{U}_{k+1}$ and $\tilde{X}_{k+1}$ are given by the SVD of $G V_{k+1}^{T}+\mu_{k}(D_{1})_{k} = \tilde{U}_{k+1}\tilde{\Sigma}_{k+1}\tilde{X}_{k+1}^{T}$. Then similarly to the results shown in \cite{ACHA:ConverageDDTF}, we can easily deduce the following conclusion.

\begin{theorem}\label{the2}
The sequence $\{D^{k}, \tilde{V}^{k}\}$ generated by (\ref{equ3.15}) converges to a stationary point of (\ref{equ3.14}).
\end{theorem}

In fact, the lack of sequence convergence is not crucial for the application of image denoising due to the reason that the results we are seeking for are not the frame coefficients but the image synthesized from the coefficients. Numerical experiments in \cite{ACHA:ConverageDDTF} also demonstrate that the algorithms with and without the proximal term shown in (\ref{equ3.15}) have almost the same denoising performance in terms of PSNR values. The conclusion also exists for the proposed algorithm. Therefore, we choose $\lambda_{k}=\mu_{k}=0$ in the following experiments.

\section{Numerical experiments}\label{sec4}

In this section, we compare the proposed algorithm (shown as Algorithm 1) with the original data-driven tight frame construction scheme \cite{ACHA:DDTF} and the recently proposed state-of-the-art models \cite{TIP:LSSC,CVPR:WNNM} in the field of AWGN denoising. Both the quality of the recovery images and the computational costs of these algorithms are compared.

\begin{figure}
  \centering
  \subfigure[]{
    \label{fig4.3:subfig:a} 
    \includegraphics[width=1.3in,clip]{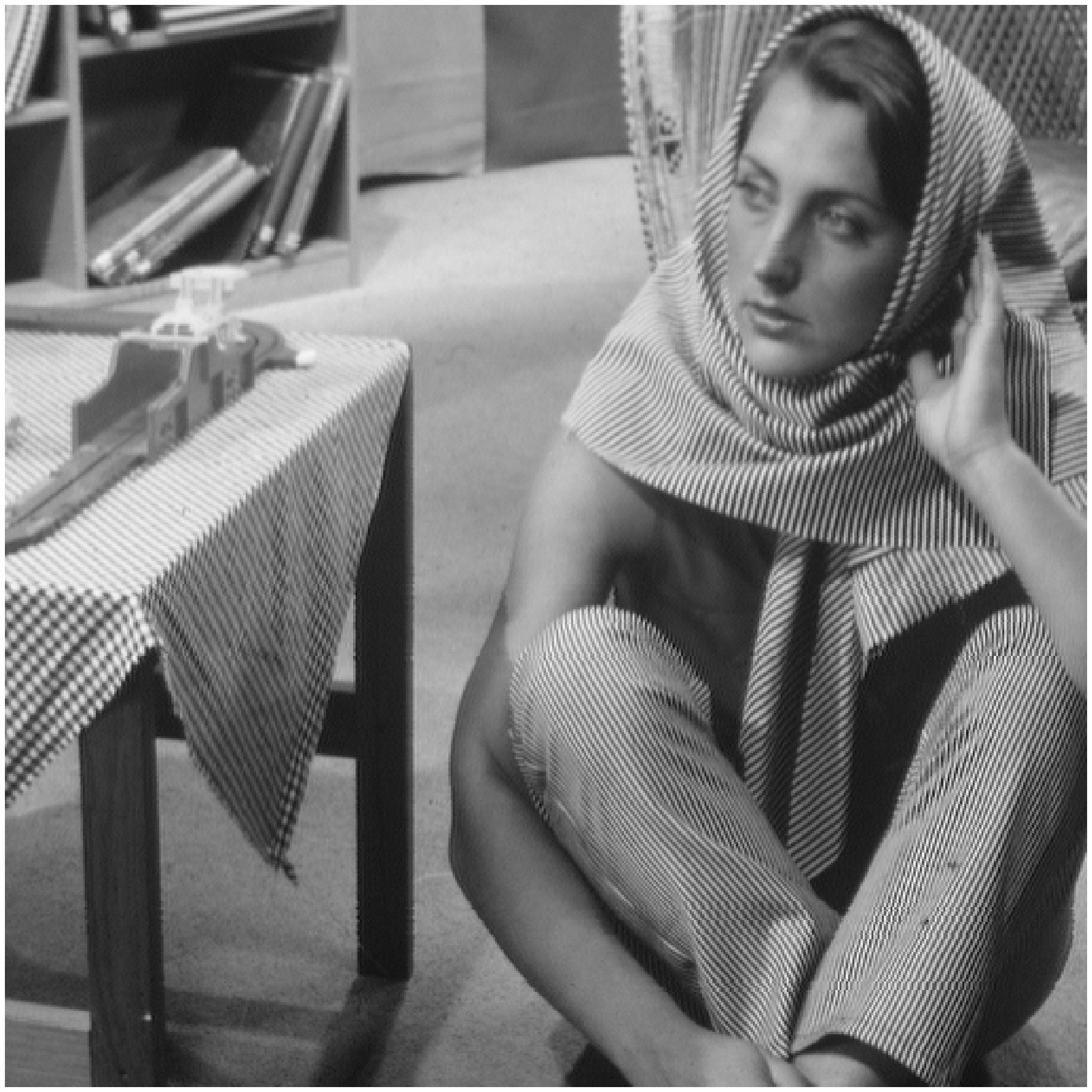}}
  \hspace{0pt}
  \subfigure[]{
    \label{fig4.3:subfig:b} 
    \includegraphics[width=1.3in,clip]{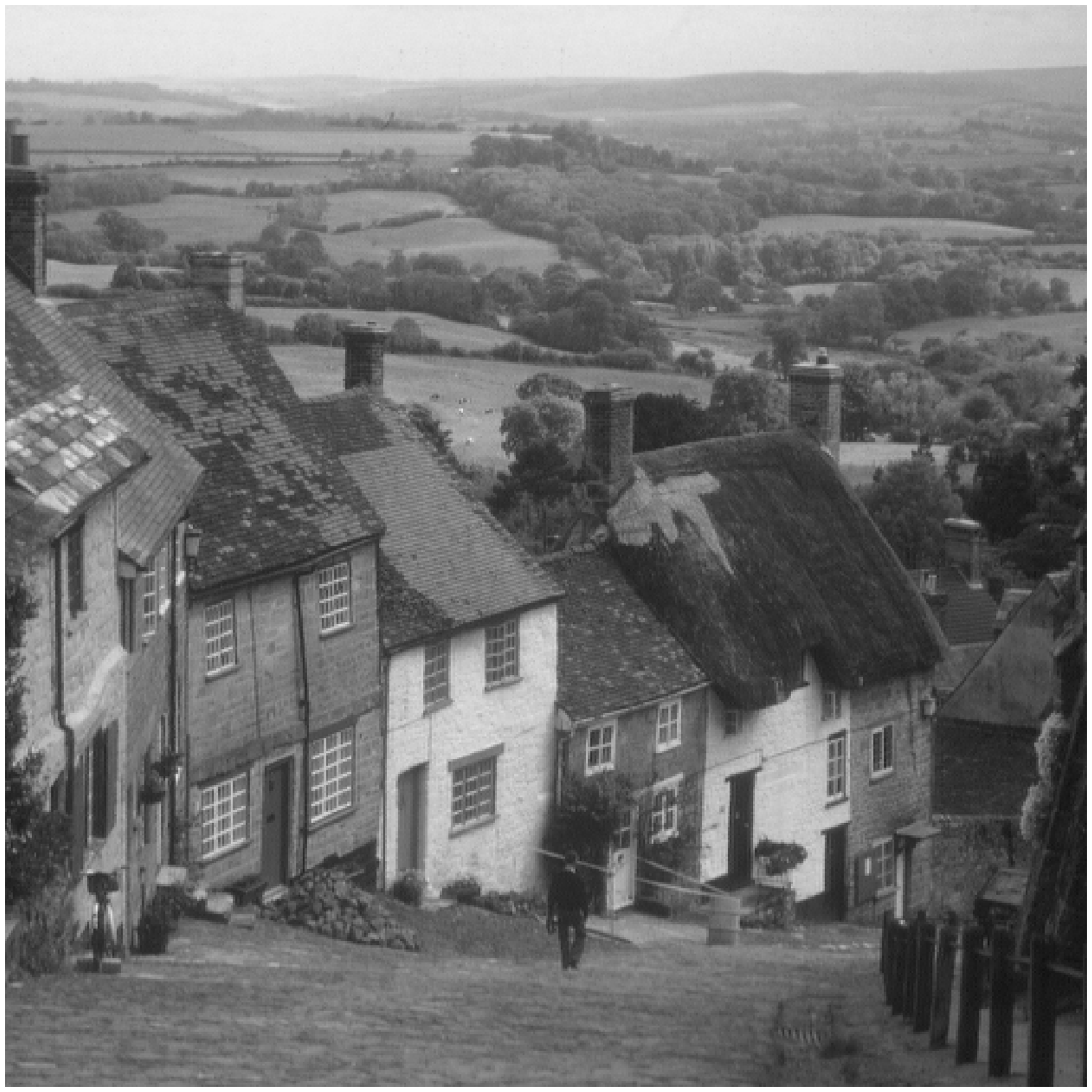}}
  \subfigure[]{
    \label{fig4.3:subfig:c} 
    \includegraphics[width=1.3in,clip]{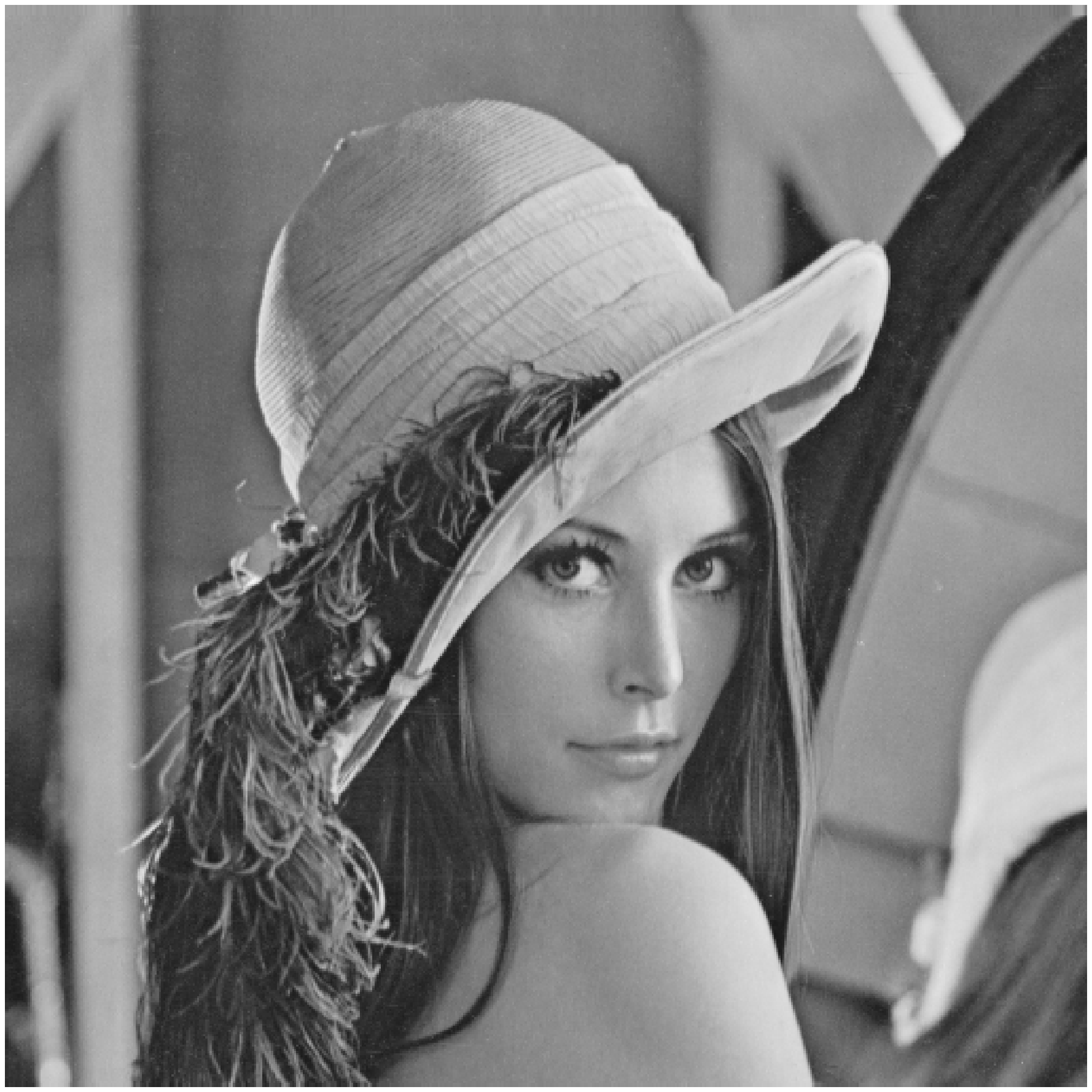}}
  \hspace{0pt}
  \subfigure[]{
    \label{fig4.3:subfig:d} 
    \includegraphics[width=1.3in,clip]{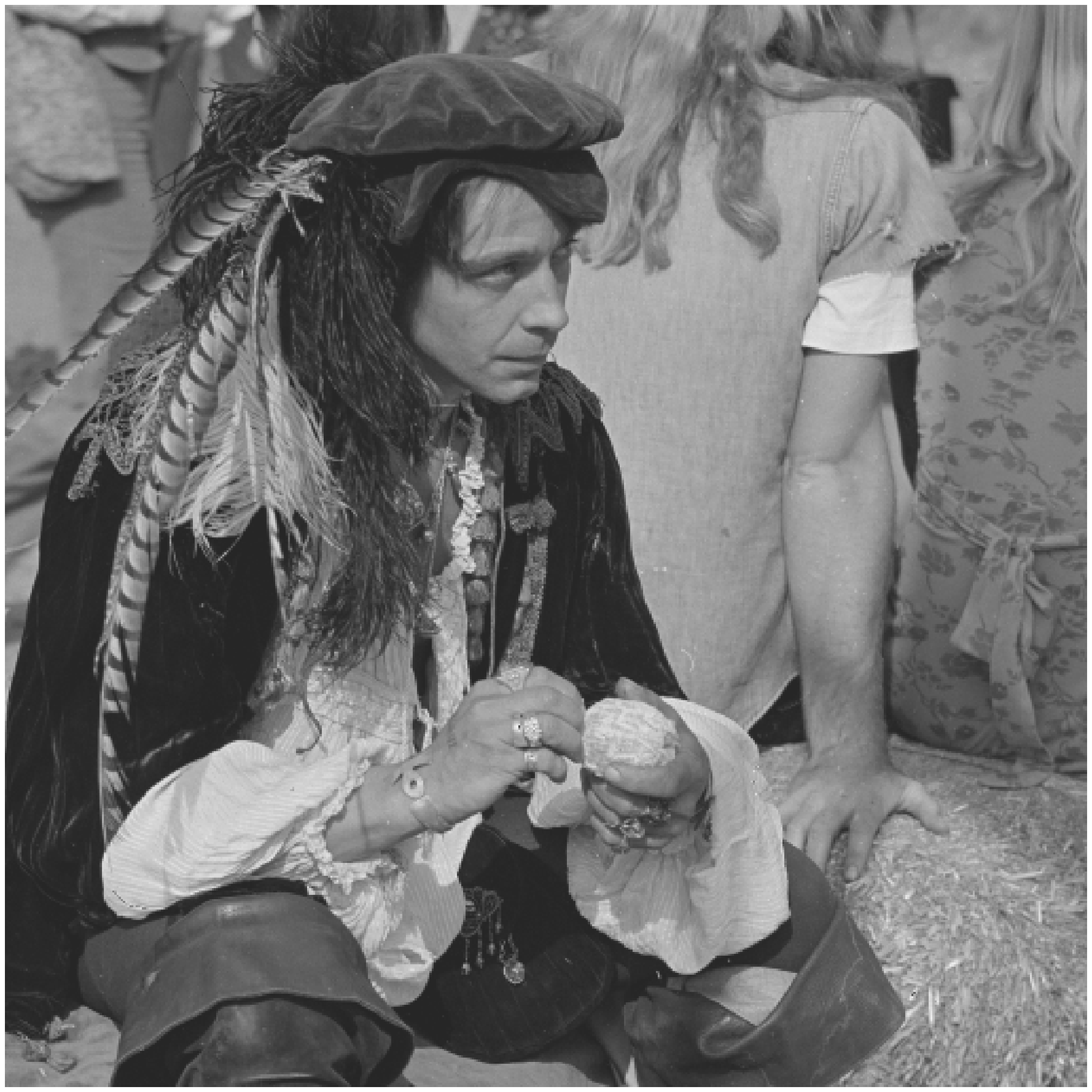}}
  \hspace{0pt}
  \subfigure[]{
    \label{fig4.3:subfig:e} 
    \includegraphics[width=1.3in,clip]{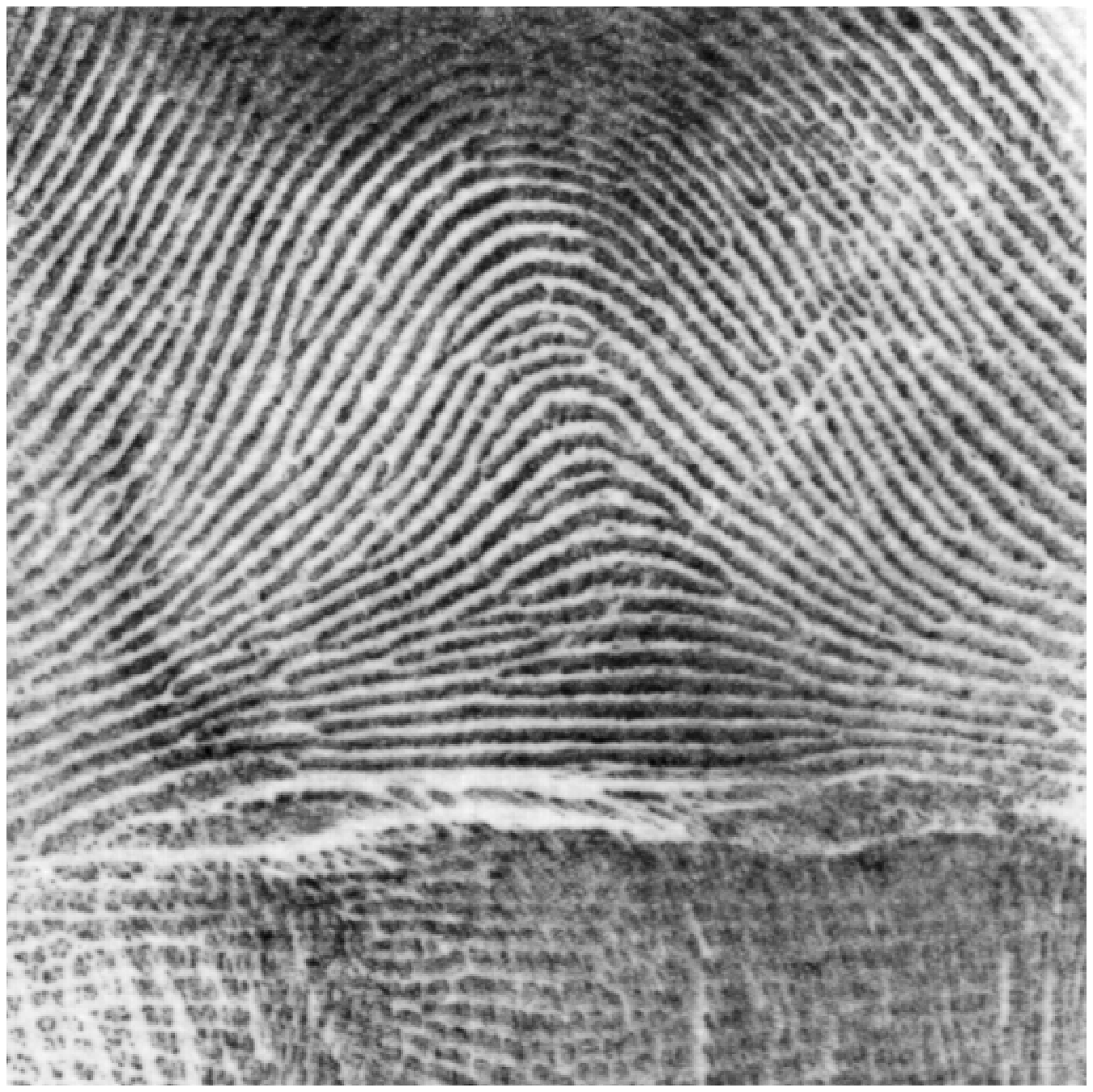}}
  \hspace{0pt}
  \subfigure[]{
    \label{fig4.3:subfig:f} 
    \includegraphics[width=1.3in,clip]{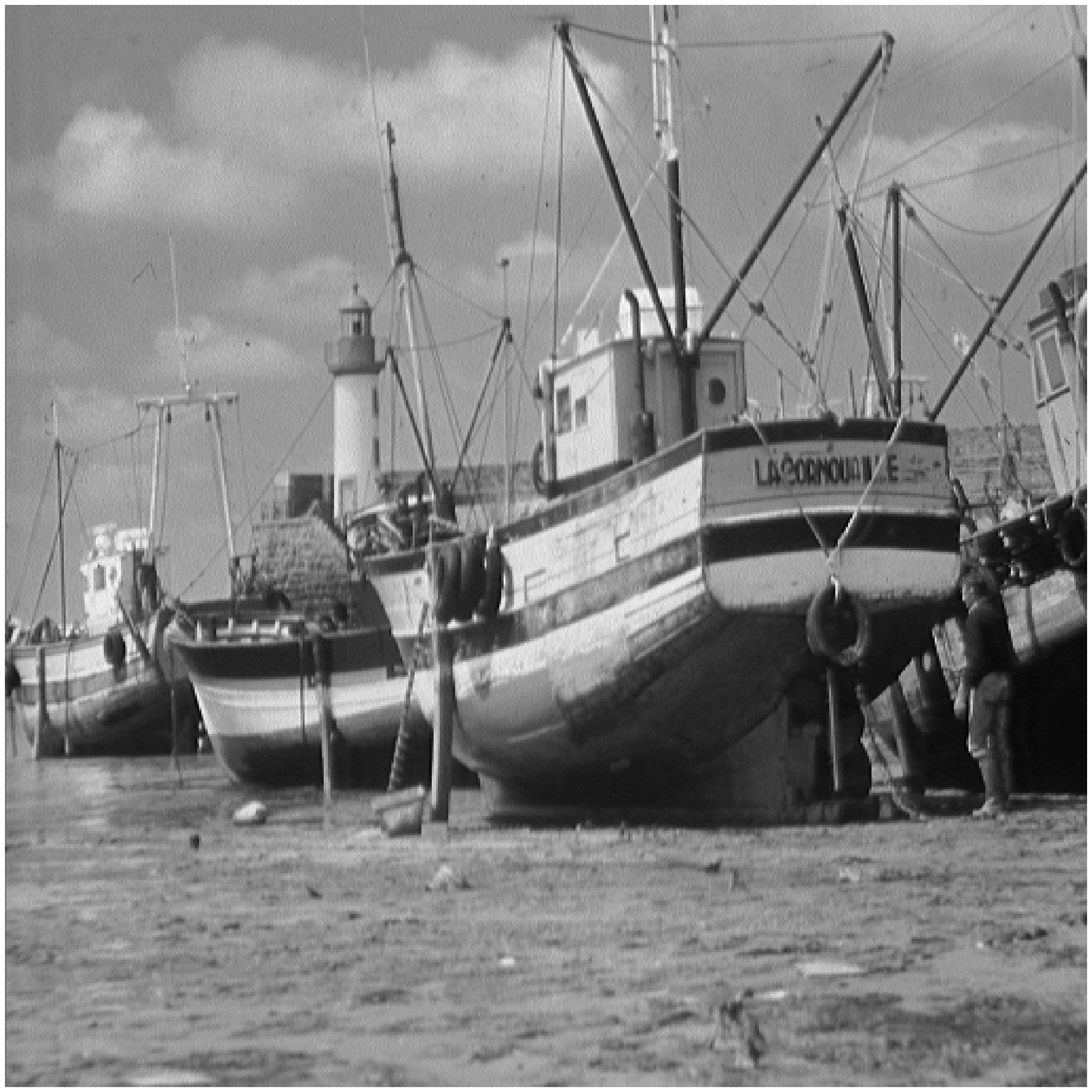}}
\caption{Original images. (a) Barbara, (b) Hill, (c) Lena, (d) Man, (e) Fingerprint, (f) Boat.
}
\label{fig4.3}
\end{figure}

The codes of Algorithm 1 and the original data-driven tight frame method \footnote{http://www.math.nus.edu.sg/~matjh/download/data \\ \_driven\_tight\_frames/data\_driven\_tight\_frame\_ver1.0.zip} are written entirely in Matlab. All the numerical examples are implemented under Windows XP and MATLAB 2009 running on a laptop with an Intel Core i5 CPU (2.8 GHz) and 8 GB Memory. In the following experiments, six standard nature images with size of $512\times 512$ (see Figure \ref{fig4.3}), which consist of complex components in different scales and with different patterns, are used for our test.

\subsection{The analysis of the proposed algorithm}\label{subsec4.1}

First of all, we simply analyze the computational costs of both the original data-driven tight frame construction scheme \cite{ACHA:DDTF} and the proposed algorithm (shown as Algorithm $1$). It is observed that each iteration of both algorithms includes two stages: hard thresholding and tight frame update. For the original algorithm in \cite{ACHA:DDTF}, the size of the matrix $D^{T}G$ used for hard thresholding operator is $p^{2}\times N$, and the tight frame update is done via the SVD of the matrix $GV^{T}$ with the size of $p^{2}\times p^{2}$. For Algorithm $1$, the size of the corresponding matrix used for hard thresholding operator is $s\times N$, and the SVD used for the tight frame update is related to the matrix with the size of $s\times p^{2}$. Therefore, the whole computational cost can be reduced obviously by choosing a small $s$. Besides, the computational burden of the analysis operator and the synthesis operator generated by $D_{1}\in \mathbb{R}^{p^{2}\times s}$, which corresponds to $s$ times of convolution operator, is also small than those produced by $D\in \mathbb{R}^{p^{2}\times p^{2}}$. Hence, the computational cost of generating the denoised image (see the formula (\ref{equ3.11})) is also smaller than that of the original data-driven tight frame method.

In what follows, the initialization of the data-driven filters set $D_{1}$ and the selection of the parameter $s$ are further discussed. The initialization of tight frames has been investigated in literature \cite{ACHA:DDTF}, and one suitable selection is to use DCT as the initial guess. Denote $\textbf{D}=\{d_{1}, d_{2}, \cdots, d_{p^{2}}\}$ as the tight frame generated by $p\times p$ DCT. We can simply generate the initial $D_{1}$ by randomly choosing $s$ elements from $\textbf{D}$. On the other hand, inspired by the discussion in section \ref{subsec3.1} (refer to (\ref{equ3.1}) and Figure \ref{fig3.1}), we can use the signal component $\textbf{S}=\sum \limits_{i=1}^{s}u_{k}^{i}(x_{k}^{i})^{T}$ to obtain the initial $D_{1}$. Specifically, we choose $s$ elements from $\textbf{S}$ that satisfy $\|e_{i}\|_{2}>\epsilon$, where $e_{i}$ denote the elements of $\textbf{S}$, and $\epsilon$ is a small constant such as $10^{-5}$ to guarantee the chosen elements are not too small and may represent the main feature of the image. The matrix $\textbf{S}$ can be generated by one or two iteration steps (In our experiments we choose two iteration steps) of the original data-driven tight frame construction scheme.

It is noted that one important parameter of the proposed algorithm is the filters number $s$. In the next, three images, Barbara, Lena and Boat, are used for testing the performance of Algorithm $1$ with different values of $s$ and initialization of filters set $D_{1}$. The size $p$ is chosen to be $8$, which is a suitable selection by considering both the recovery quality and implementation time. For the thresholding parameters $\lambda$ and $\tilde{\lambda}$, we use the default setting in the recent published code of literature \cite{ACHA:DDTF}, i.e., fix $\lambda=3.4 \sigma$ and $\tilde{\lambda}=2.7 \sigma$, which is the suitable selection through many trials. The maximum iteration number is set to $25$.
The performance is quantitatively measured by means of PSNR, which is expressed as

\begin{equation}\label{equ4.1}
\textrm{PSNR}(u,u^{*})=-20\lg
\left\{\frac{\|u-u^{*}\|_2}{255 N}\right\}
\end{equation}
where $u$ and $u^{*}$ denote the original and restored images, respectively, and $N$ is the total number of pixels in the image $u$.

Table \ref{tab4.1} lists the PSNR values and CPU time for various selection of $s$ values and initialization of $D_{1}$. Here $s=10, \cdots, 50$ represent that the corresponding initialization of $D_{1}$ is obtained by the signal component $\textbf{S}$, and $DCT, s=10, 20$ demonstrate that the initialization is given by the DCT.
It is observed that $s=20$ or $30$ is a suitable selection for Algorithm $1$, which means that $s$ can be chosen as $p^2/3$ or $p^2/2$ approximately. Besides, the performance of the algorithm with the initialization of $D_{1}$ from $\textbf{S}$ overall outperforms that using the DCT as the initialization. In the following experiments, we use the better initialization strategy for $D_{1}$.

\begin{table*} [htbp]
\centering \caption{The comparison of the performance of Algorithm $1$ with different $s$ and initialization}
\scalebox{0.9}{
\begin{tabular}{cccccccccc}
  \hline
  Image & noise & index & $s=10$ &  $s=20$ & $s=30$ & $s=40$ & $s=50$ & $DCT, s=20$ & $DCT, s=30$\\
  \hline
  \multirow{6}{*}{Barbara} & \multirow{2}{*}{$\sigma=30$} & PSNR & 26.16 & 27.84 & \textbf{28.57} & 28.56 & 28.53 & 27.83 & 28.54\\
  \cline{3-10}
   &  & Time & 1.10 & 1.62 & 2.21 & 2.79 & 3.39 & 1.59 & 2.15\\
  \cline{2-10}
  &  \multirow{2}{*}{$\sigma=40$} & PSNR & 25.21 & 26.54 & \textbf{27.15} & 27.12 & 27.07 & 26.69 & 27.08 \\
  \cline{3-10}
  &  & Time & 1.06 & 1.68 & 2.16 & 2.73 & 3.43 & 1.57 & 2.15 \\
  \cline{2-10}
  &  \multirow{2}{*}{$\sigma=50$} & PSNR & 24.94 & 25.64 & \textbf{26.00} & 25.89 & 25.88 & 25.54 & 25.80 \\
  \cline{3-10}
  &  & Time & 1.09 & 1.63 & 2.19 & 2.73 & 3.26 & 1.73 & 2.11 \\
  \hline
  \hline
  \multirow{6}{*}{Lena} & \multirow{2}{*}{$\sigma=30$} & PSNR & 30.41 & 30.58 & \textbf{30.59} & 30.51 & 30.44 & 30.52 & 30.51\\
  \cline{3-10}
   &  & Time & 1.09 & 1.68 & 2.17 & 2.72 & 3.29 & 1.58 & 2.14\\
  \cline{2-10}
  &  \multirow{2}{*}{$\sigma=40$} & PSNR & 29.18 & \textbf{29.23} & 29.18 & 29.12 & 29.02 & 29.08 & 29.14 \\
  \cline{3-10}
  &  & Time & 1.07 & 1.64 & 2.19 & 2.72 & 3.20 & 1.56 & 2.13 \\
  \cline{2-10}
  &  \multirow{2}{*}{$\sigma=50$} & PSNR & 28.15 & \textbf{28.16} & 28.07 & 27.99 & 27.86 & 28.06 & 28.04 \\
  \cline{3-10}
  &  & Time & 1.10 & 1.63 & 2.20 & 2.74 & 3.29 & 1.57 & 2.12 \\
  \hline \hline
  \multirow{6}{*}{Boat} & \multirow{2}{*}{$\sigma=30$} & PSNR & 28.01 & 28.48 & \textbf{28.53} & 28.52 & 28.47 & 28.51 & 28.48\\
  \cline{3-10}
   &  & Time & 1.07 & 1.63 & 2.18 & 2.72 & 3.31 & 1.57 & 2.11\\
  \cline{2-10}
  &  \multirow{2}{*}{$\sigma=40$} & PSNR & 26.98 & \textbf{27.21} & \textbf{27.21} & 27.20 & 27.15 & 27.15 & 27.14 \\
  \cline{3-10}
  &  & Time & 1.06 & 1.65 & 2.19 & 2.70 & 3.26 & 1.72 & 2.12 \\
  \cline{2-10}
  &  \multirow{2}{*}{$\sigma=50$} & PSNR & 26.11 & \textbf{26.24} & 26.17 & 26.15 & 26.08 & 26.14 & 26.16 \\
  \cline{3-10}
  &  & Time & 1.08 & 1.64 & 2.15 & 2.73 & 3.27 & 1.56 & 2.13 \\
  \hline
\end{tabular}}
\label{tab4.1}
\end{table*}

\subsection{Denoising performance evaluation on additive white Gaussian noise}\label{subsec4.2}

In this section, we report the experimental results, comparing the proposed algorithm with the original data-driven tight frame construction scheme \cite{ACHA:DDTF}, on the denoising of additive Gaussian noise. Six natural images with size of $512\times 512$ are used for our test. Considering both the recovery quality and computational time, we set the maximum iteration number to $25$ for both algorithms through all experiments. In this experiment, we consider the filters with size of $8\times 8$ and $16\times 16$ respectively. The selection of the thresholding parameters $\lambda$ and $\tilde{\lambda}$ is the same as that mentioned above. Table \ref{tab4.2} lists the PSNR values and CPU time of different algorithms. Here the original data-driven tight frame construction scheme \cite{ACHA:DDTF} is abbreviated as ``DDTF", and ``DDTF; 8" represents the DDTF method with $p=8$. Similarly, ``Alg1(20);8" denotes Algorithm $1$ with $p=8$ and $s=20$, where the number in the bracket represents the value of $s$, i.e., the number of chosen filters. Two values of $s$, which approximately equal to $p^2/3$ or $p^2/2$, are tested here.

From the results in Table \ref{tab4.2}, we observe that the proposed algorithm overall outperforms the DDTF method in both the recovery quality and computational time, especially for the higher noise level. This is due to more filters learned from the SVD of the matrix $GV^{T}$ are influenced by the several noise. As a result, the coefficients generated by these filters may not be sparse, and removing these filters can improve the recovery quality and reduce the computational cost meanwhile. In general, comparing the indexes in Table \ref{tab4.2}, we can easily draw the following conclusions.

1) The DDTF method with $p=8$ takes about $4.0\sim 4.1 s$, Algorithm $1$ with $p=8$ and $s=20$ takes about $1.6\sim 1.7 s$, and Algorithm $1$ with $p=8$ and $s=30$ takes about $2.1\sim 2.2 s$. The computational time of the corresponding algorithms increases more than ten times due to the fact that both the size of the support set of filters and the number of filters increase four times.

2) The small $s$ is suitable for images with less textures and higher noise level, and the improvement of recovery quality becomes more obvious while noise level increases. For the case of $p=8$, it is observed that Algorithm $1$ with $s=20$ obtains the best PSNRs for almost all the condition that $\sigma\geq 40$ (except the Barbara image which is rich in terms of textures).

3) These algorithms with $p=16$ outperform those with $p=8$ for the images Barbara, Lena and Fingerprint, especially when the noise is large. Maybe it is the reason that these images have more complex texture regions.

\begin{figure}
  \centering
  \subfigure[]{
    \label{fig4.4:subfig:a} 
    \includegraphics[width=1.3in,clip]{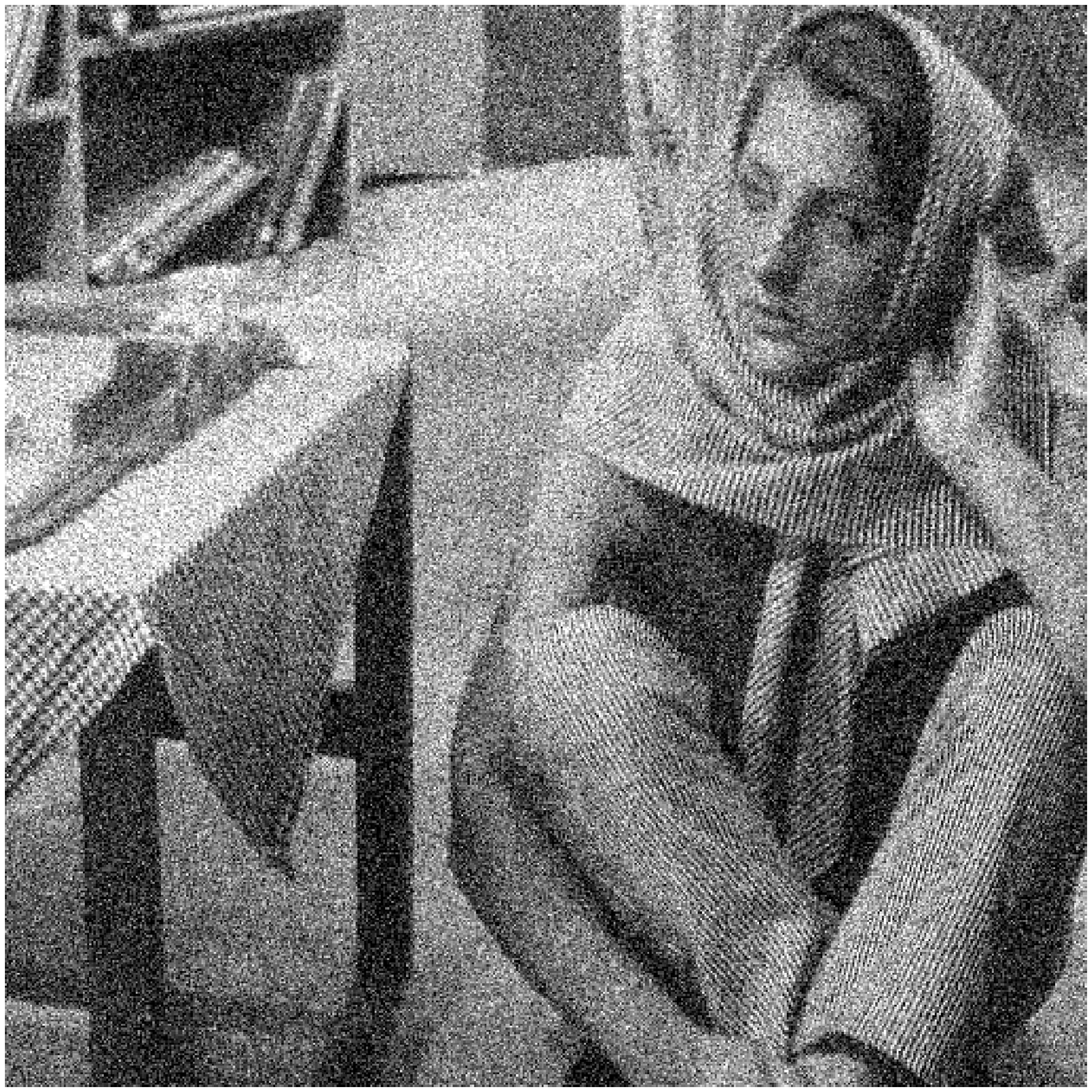}}
  \hspace{0pt}
  \subfigure[]{
    \label{fig4.4:subfig:b} 
    \includegraphics[width=1.3in,clip]{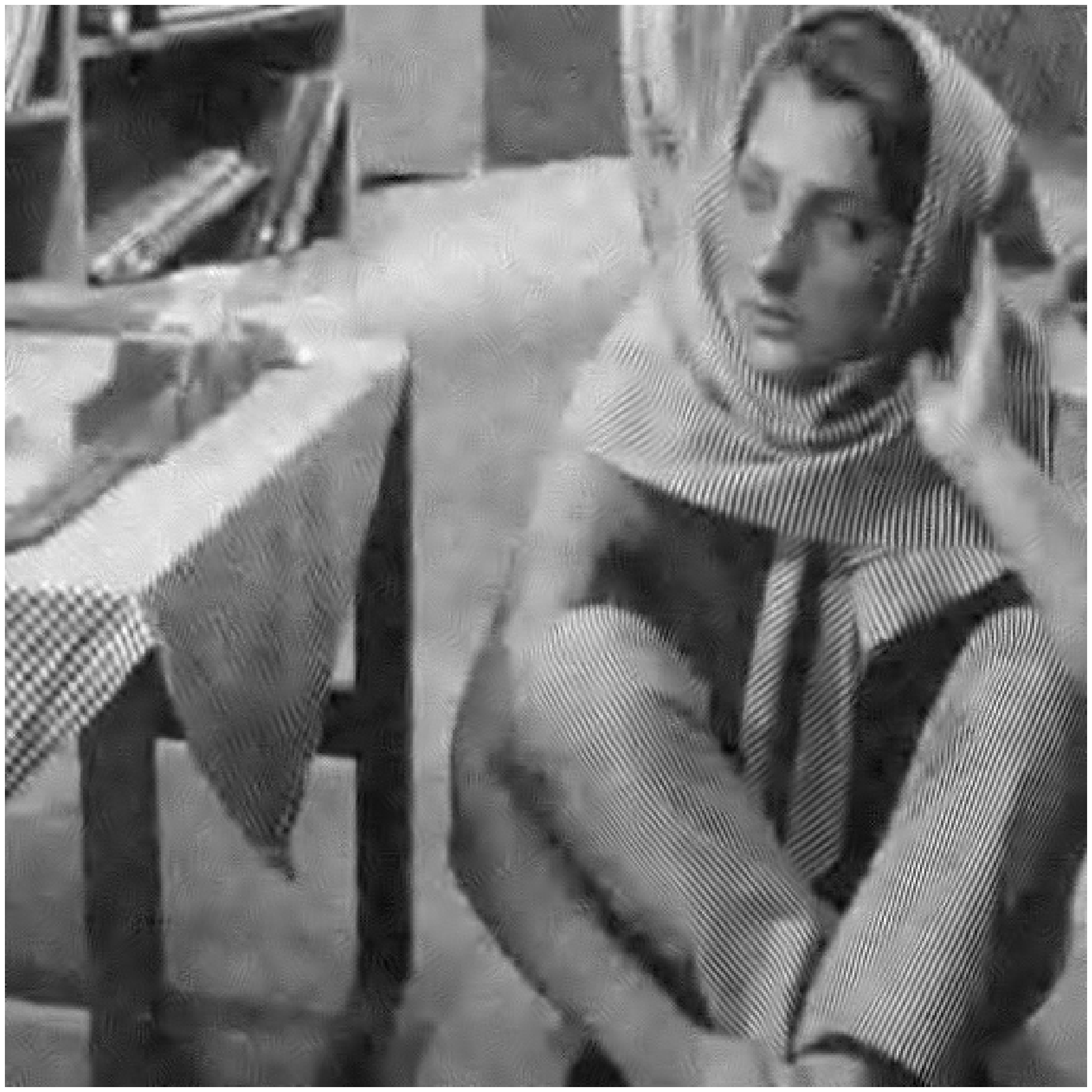}}
  \subfigure[]{
    \label{fig4.4:subfig:c} 
    \includegraphics[width=1.3in,clip]{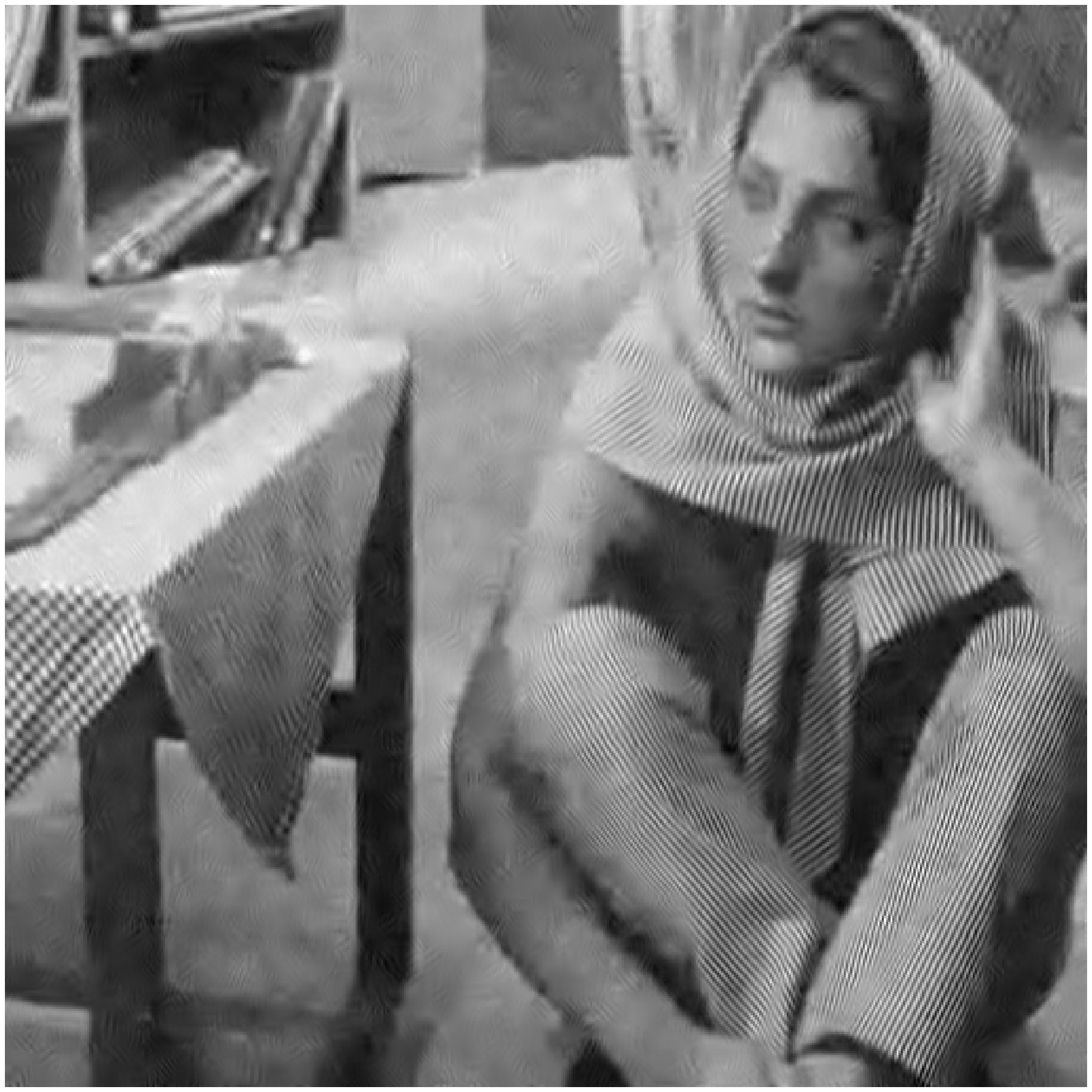}}
  \hspace{0pt}
  \subfigure[]{
    \label{fig4.4:subfig:d} 
    \includegraphics[width=1.3in,clip]{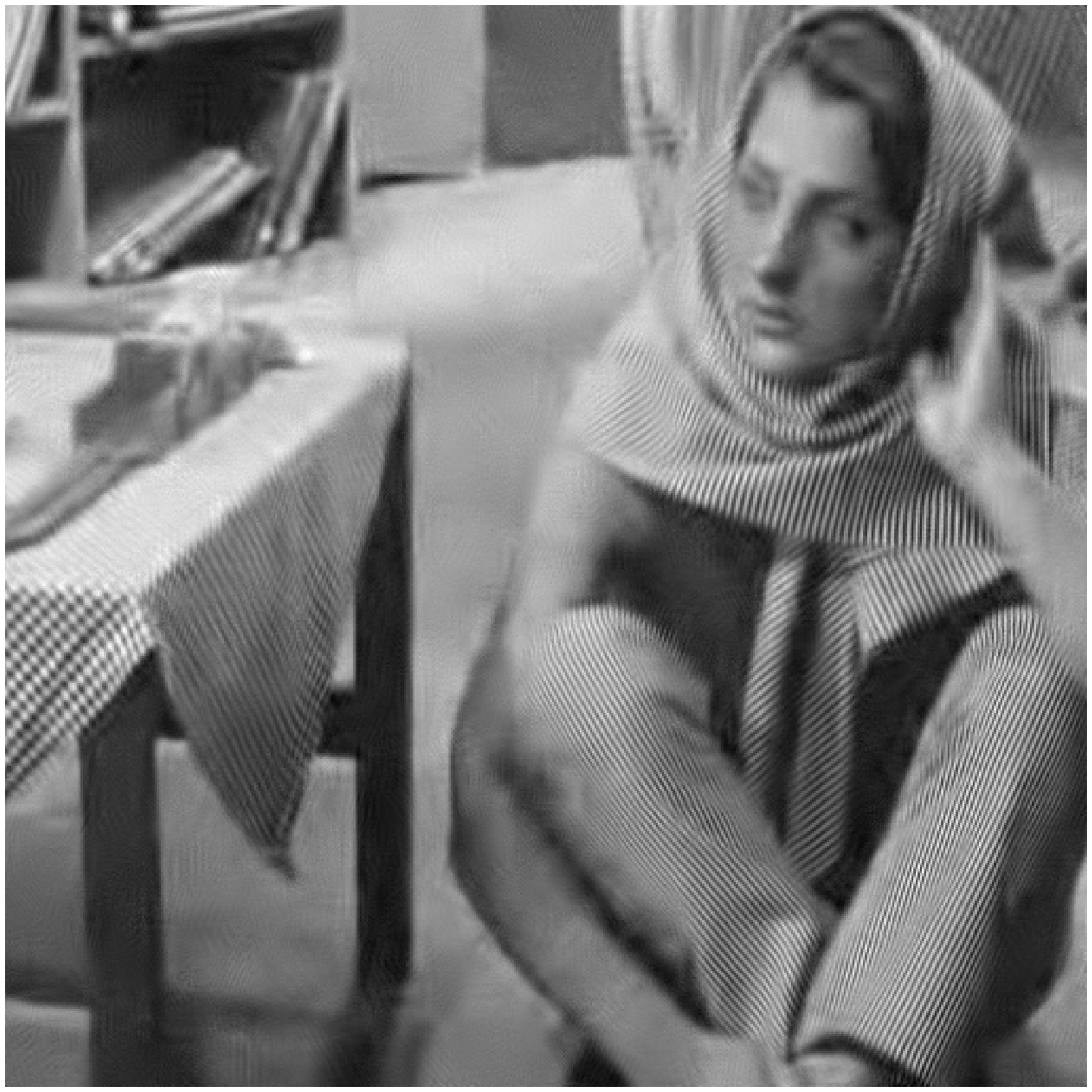}}
  \hspace{0pt}
  \subfigure[]{
    \label{fig4.4:subfig:e} 
    \includegraphics[width=1.3in,clip]{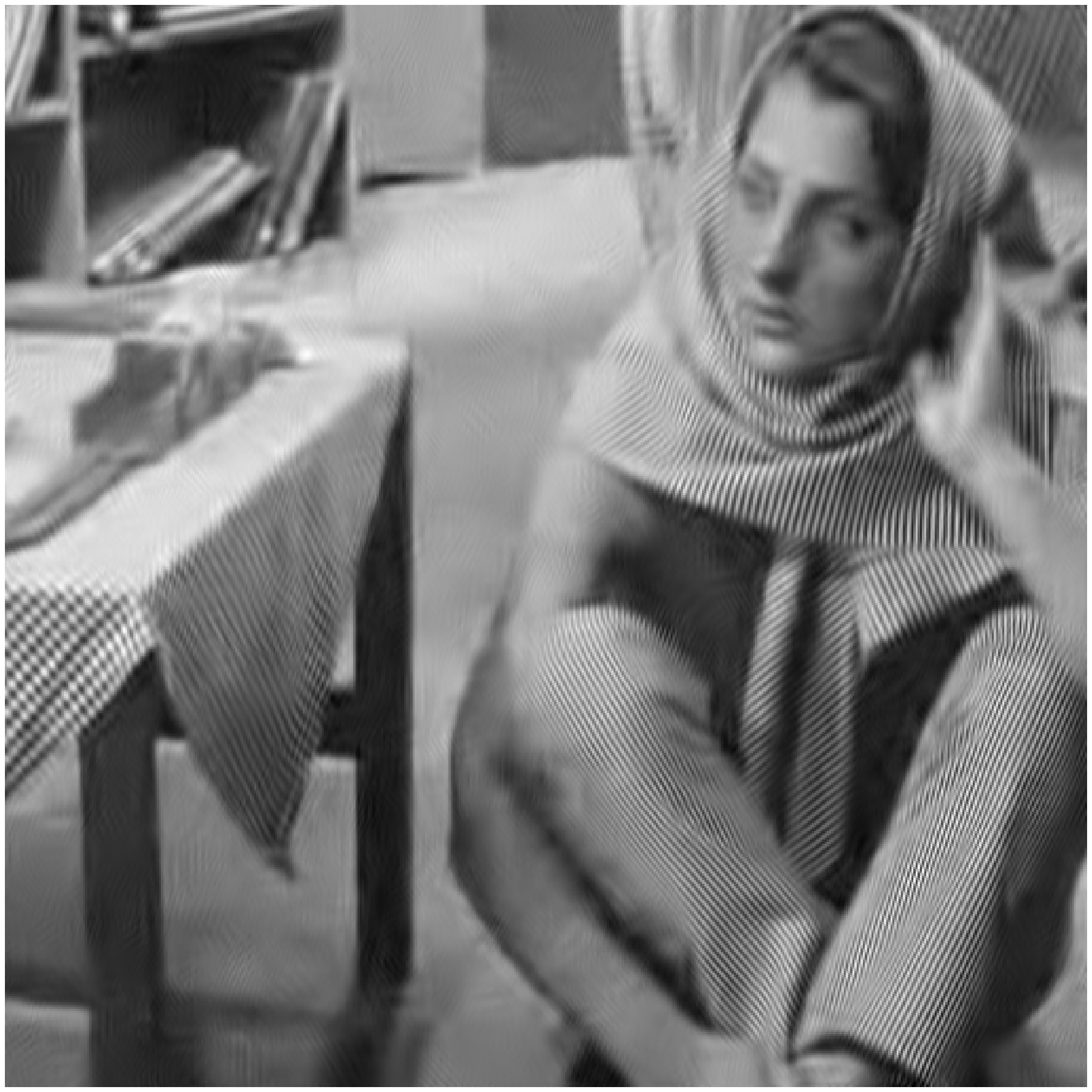}}
  \hspace{0pt}
  \subfigure[]{
    \label{fig4.4:subfig:f} 
    \includegraphics[width=1.3in,clip]{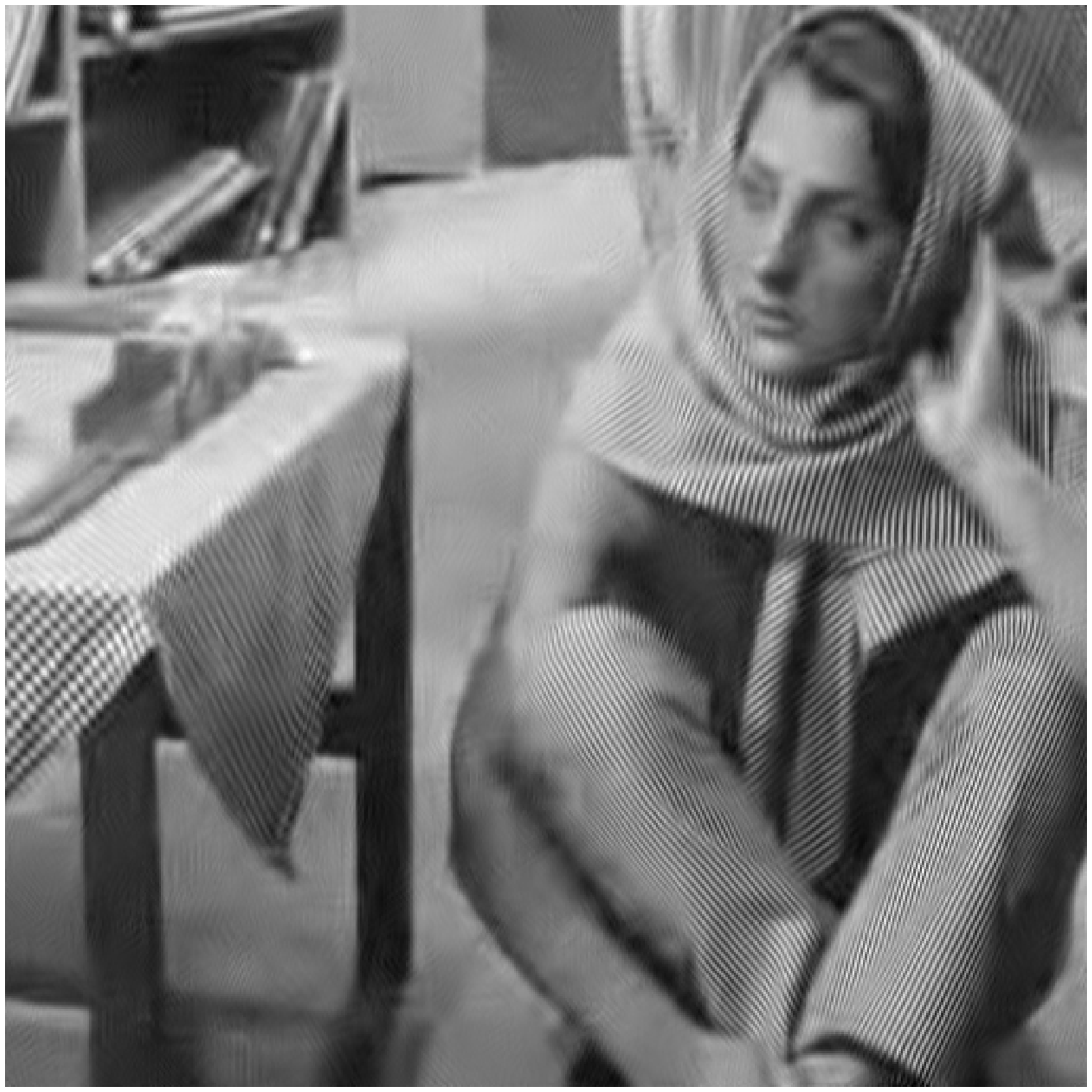}}
\caption{Visual comparison of denoising results. (a) Noisy image with $\sigma=40$, (b) result by DDTF method with $p=8$, (c) result by Alg.1 with $p=8$ and $s=30$, (d) result by DDTF method with $p=16$, (e) result by Alg.1 with $p=16$ and $s=80$, (f) result by Alg.1 with $p=16$ and $s=120$.
}
\label{fig4.4}
\end{figure}

\begin{figure}
  \centering
  \subfigure[]{
    \label{fig4.5:subfig:a} 
    \includegraphics[width=1.3in,clip]{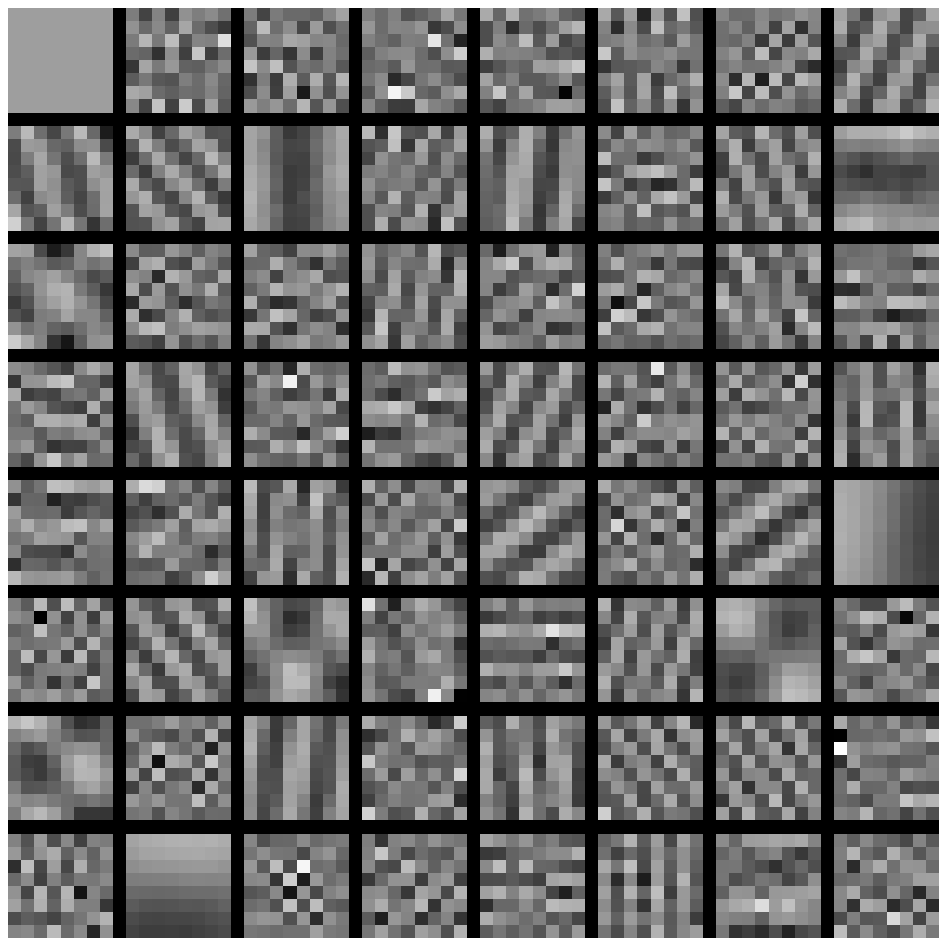}}
  \hspace{0pt}
  \subfigure[]{
    \label{fig4.5:subfig:b} 
    \includegraphics[width=1.3in,clip]{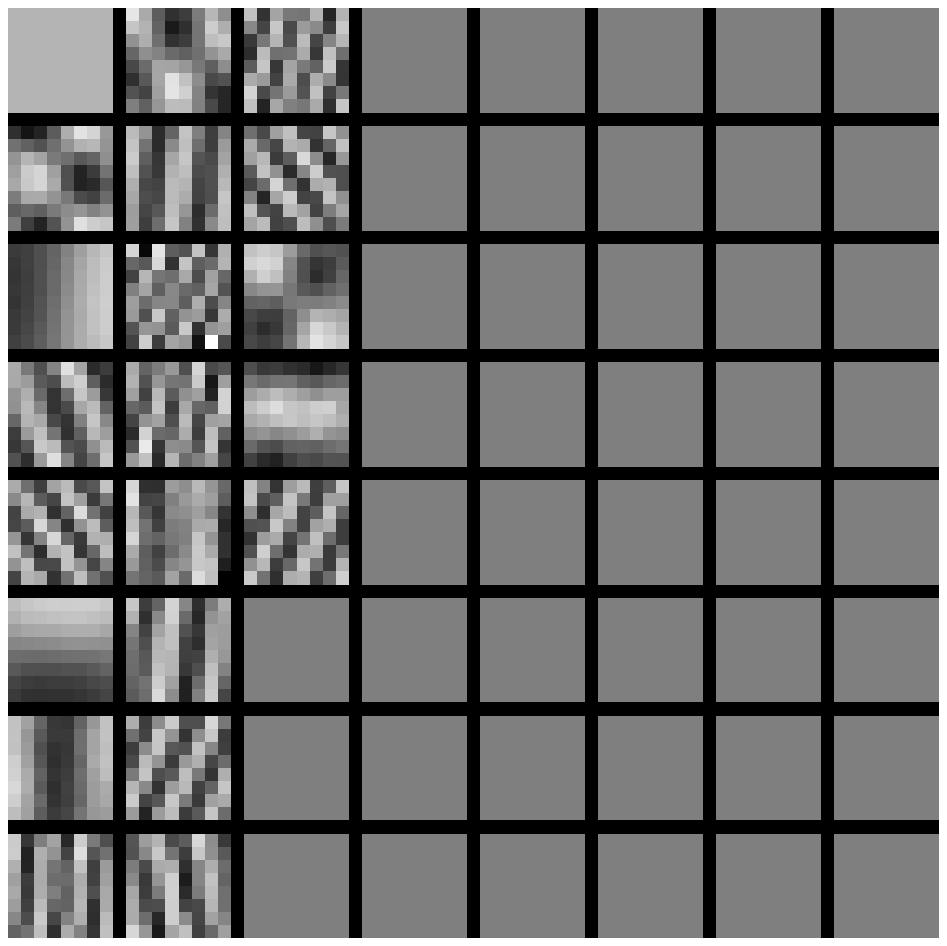}}
  \subfigure[]{
    \label{fig4.5:subfig:c} 
    \includegraphics[width=1.3in,clip]{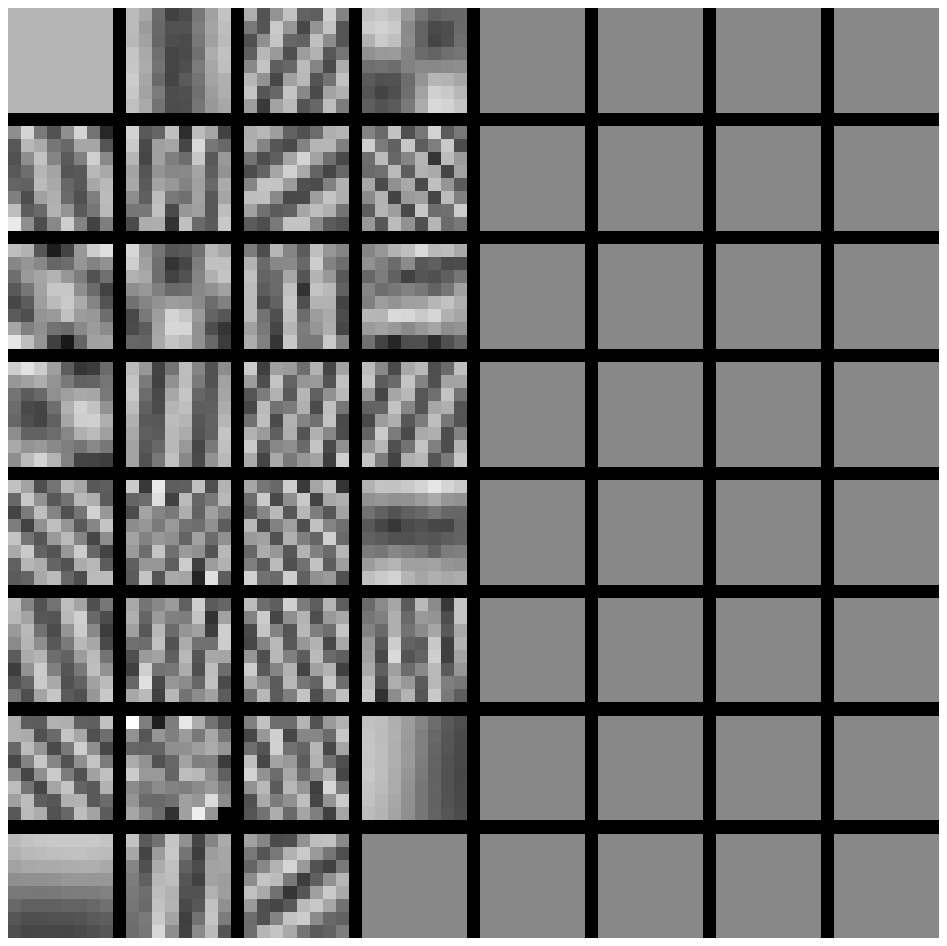}}
  \hspace{0pt}
  \subfigure[]{
    \label{fig4.5:subfig:d} 
    \includegraphics[width=1.3in,clip]{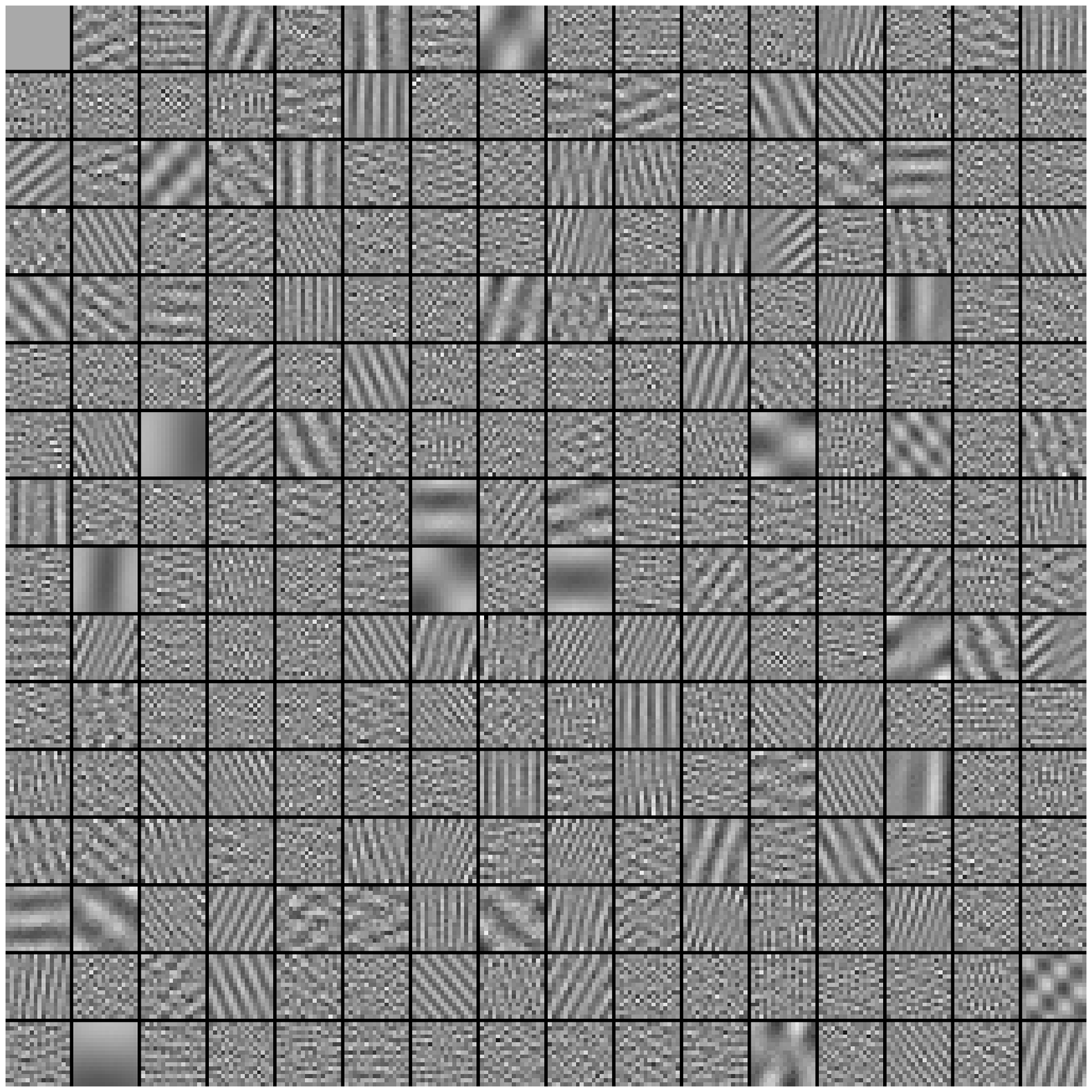}}
  \hspace{0pt}
  \subfigure[]{
    \label{fig4.5:subfig:e} 
    \includegraphics[width=1.3in,clip]{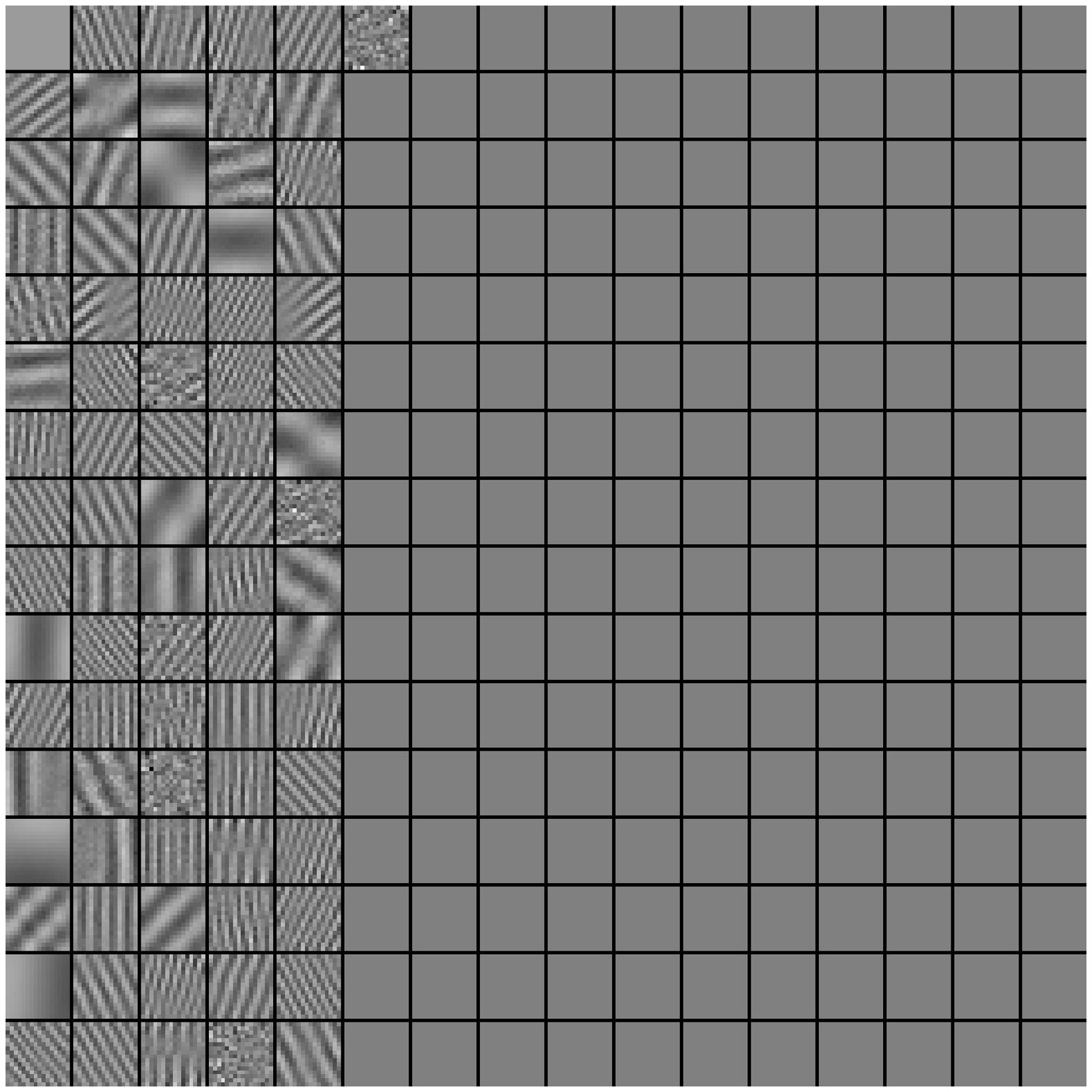}}
  \hspace{0pt}
  \subfigure[]{
    \label{fig4.5:subfig:f} 
    \includegraphics[width=1.3in,clip]{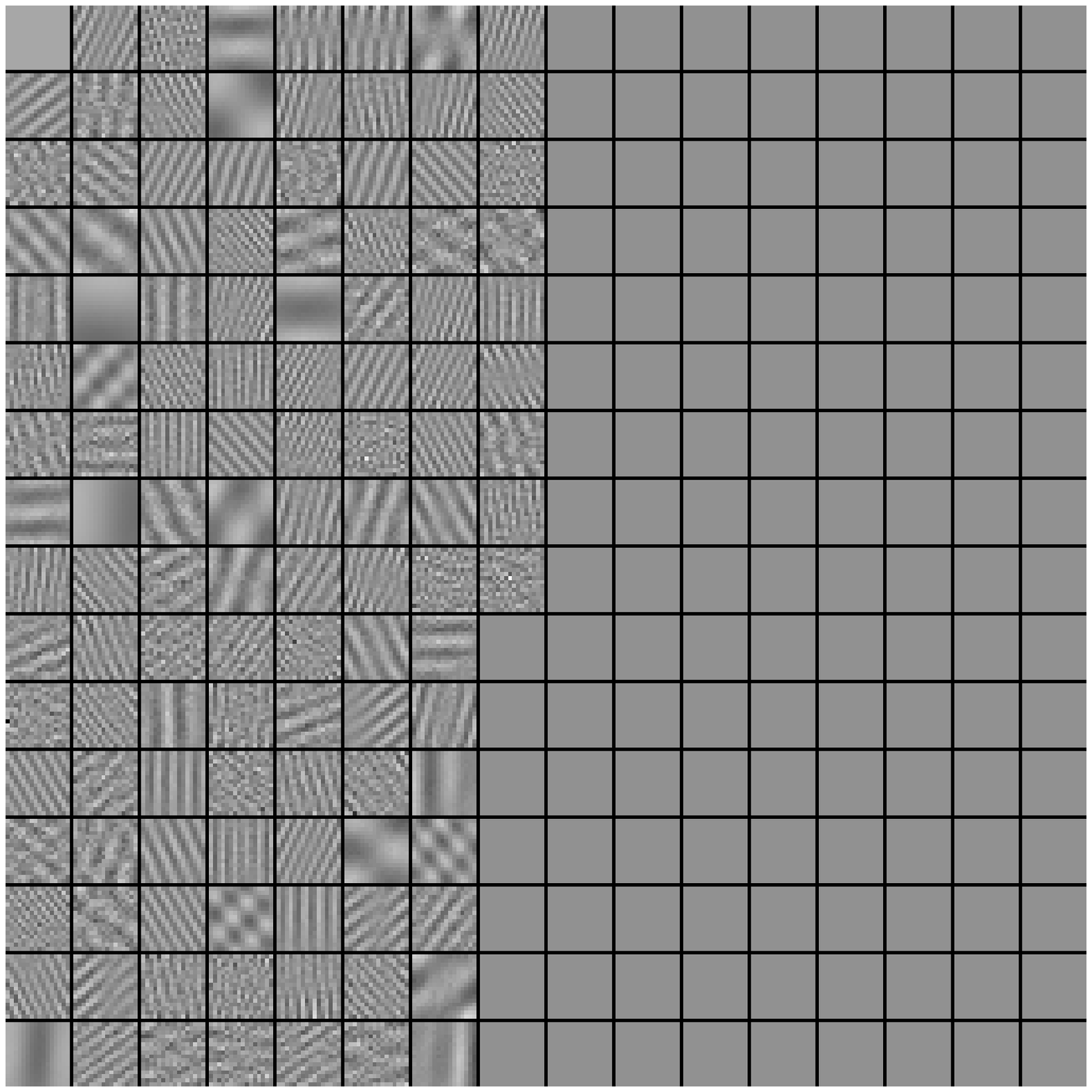}}
\caption{The data-driven tight frame filters. (a) result by DDTF method with $p=8$, (b) result by Alg.1 with $p=8$ and $s=20$, (c) result by Alg.1 with $p=8$ and $s=30$, (d) result by DDTF method with $p=16$, (e) result by Alg.1 with $p=16$ and $s=80$, (f) result by Alg.1 with $p=16$ and $s=120$.
}
\label{fig4.5}
\end{figure}

\begin{figure}
  \centering
  \subfigure[]{
    \label{fig4.6:subfig:a} 
    \includegraphics[width=1.3in,clip]{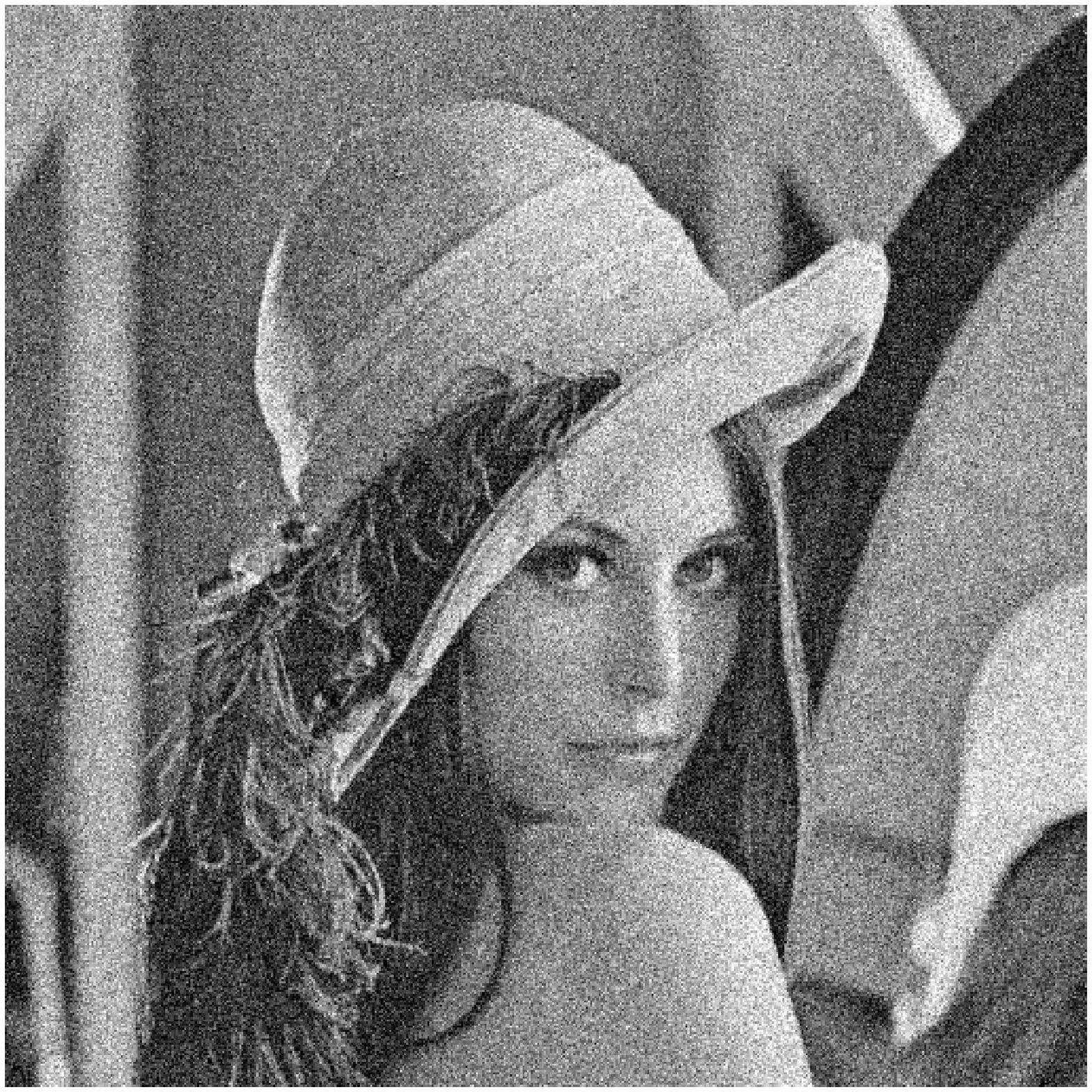}}
  \hspace{0pt}
  \subfigure[]{
    \label{fig4.6:subfig:b} 
    \includegraphics[width=1.3in,clip]{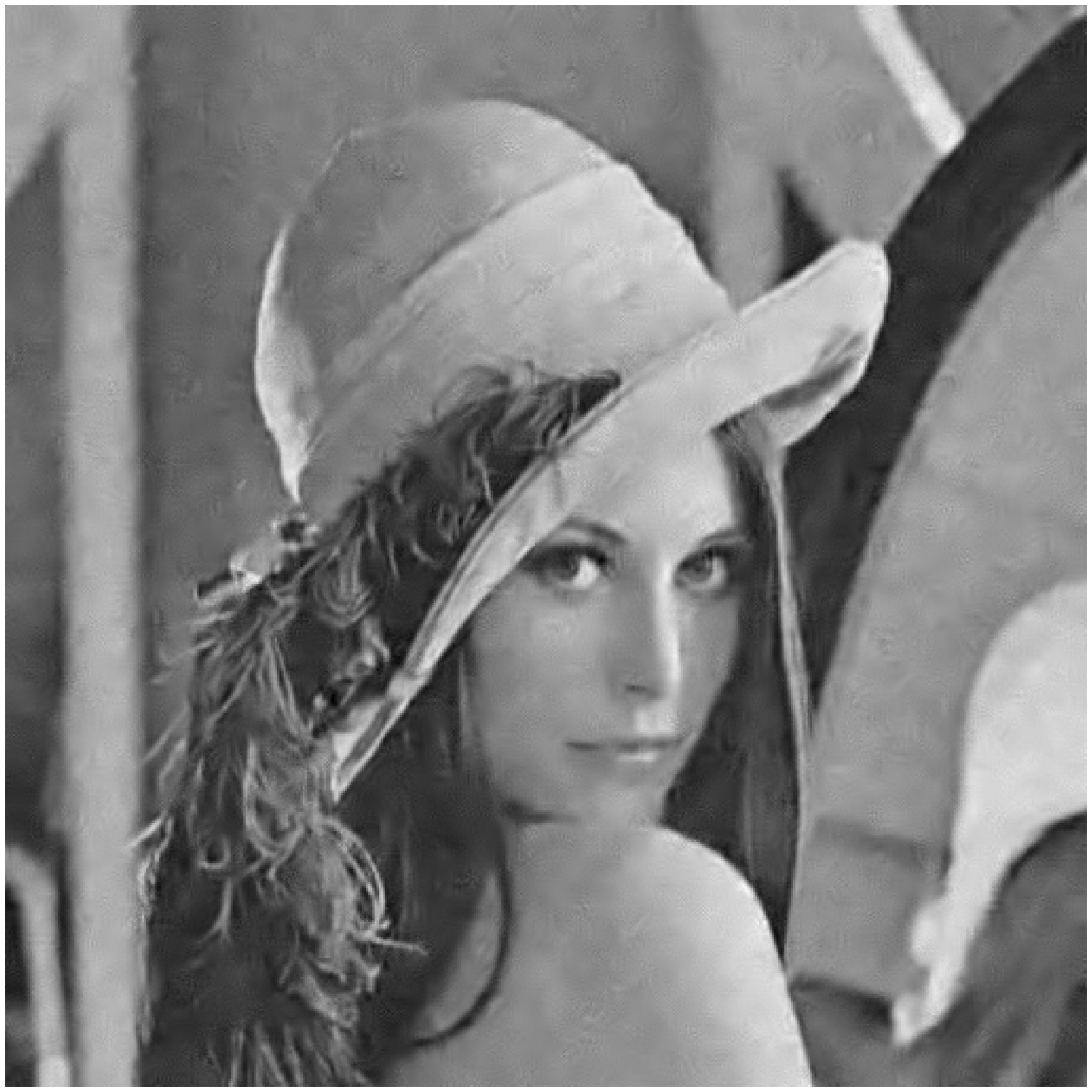}}
  \subfigure[]{
    \label{fig4.6:subfig:c} 
    \includegraphics[width=1.3in,clip]{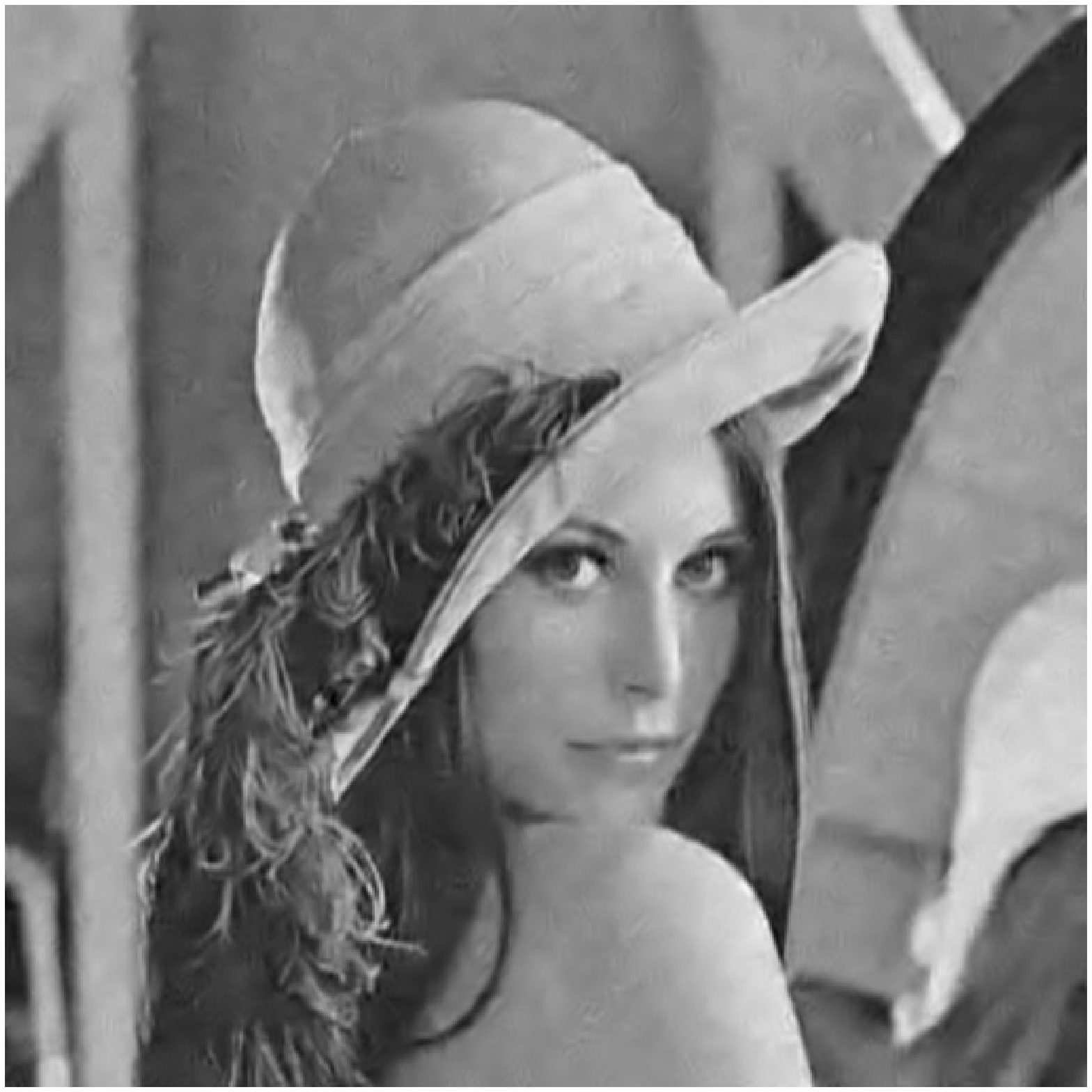}}
  \hspace{0pt}
  \subfigure[]{
    \label{fig4.6:subfig:d} 
    \includegraphics[width=1.3in,clip]{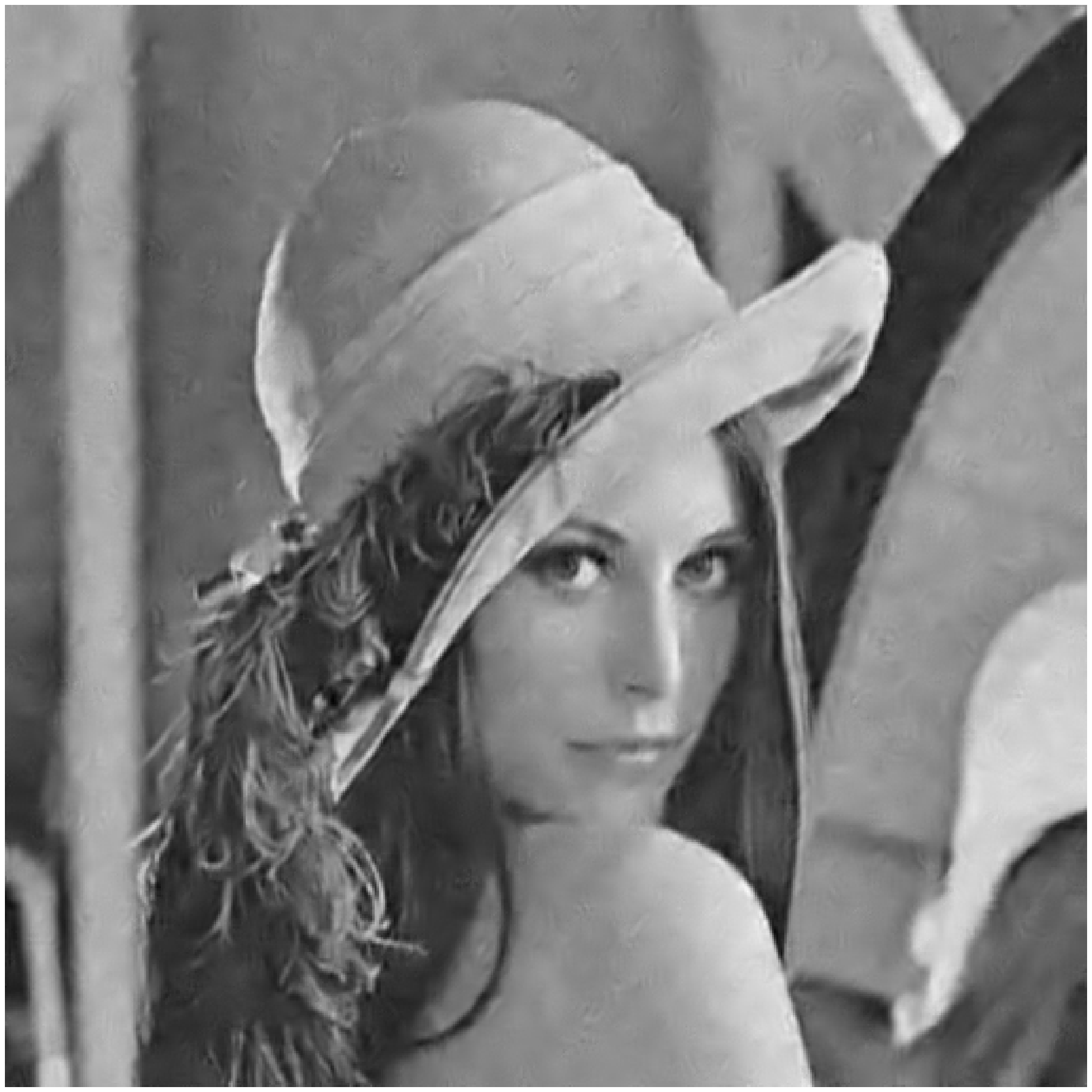}}
  \hspace{0pt}
  \subfigure[]{
    \label{fig4.6:subfig:e} 
    \includegraphics[width=1.3in,clip]{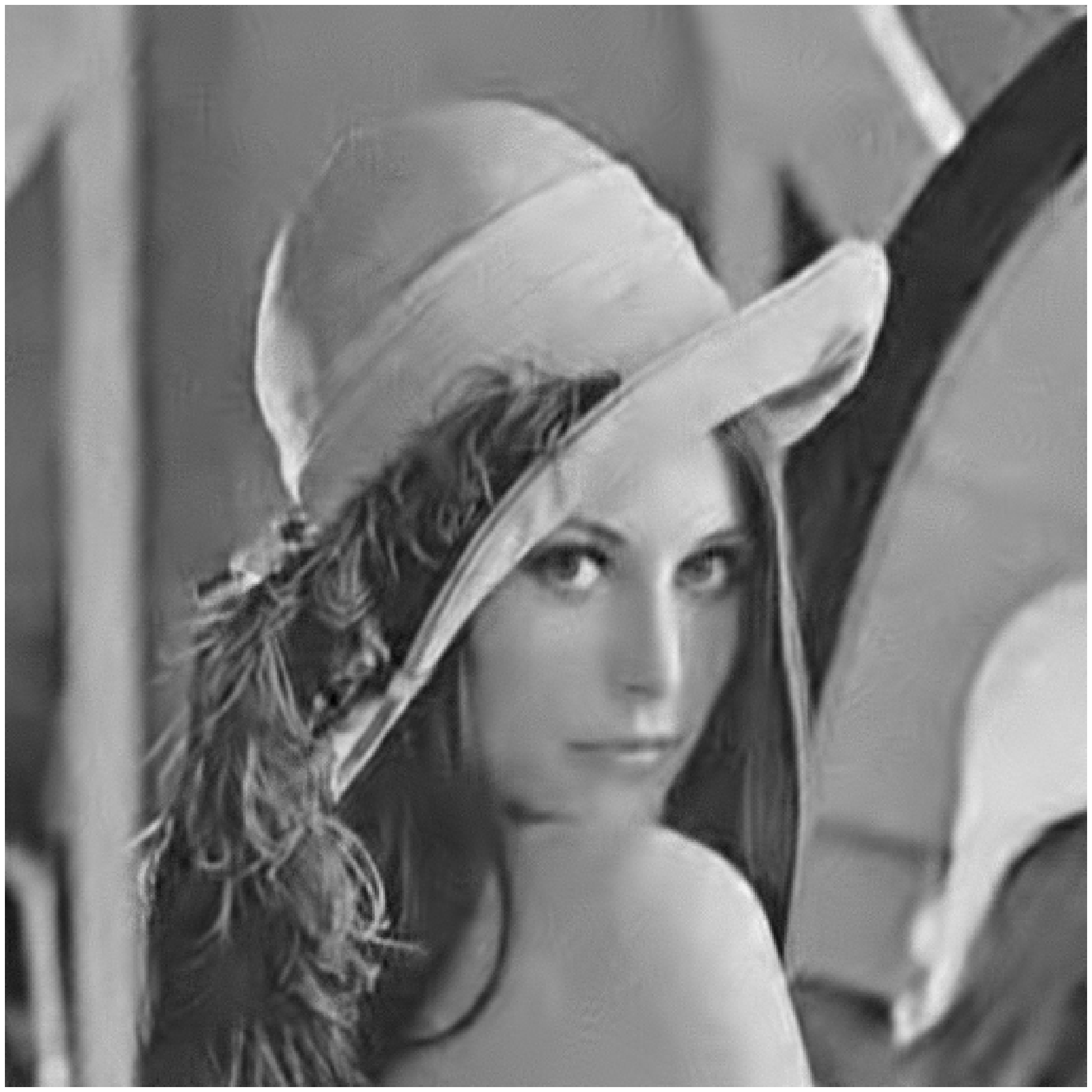}}
  \hspace{0pt}
  \subfigure[]{
    \label{fig4.6:subfig:f} 
    \includegraphics[width=1.3in,clip]{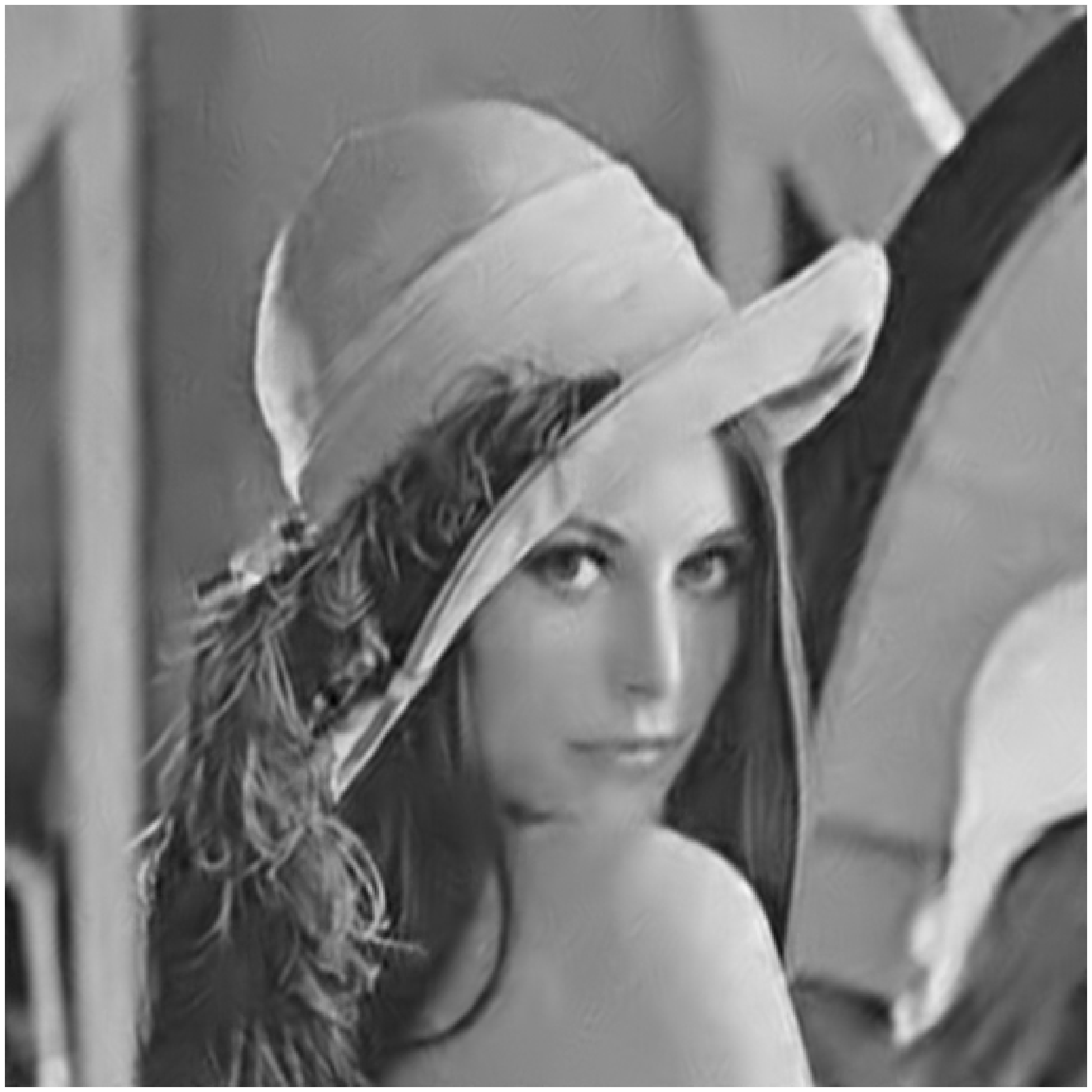}}
\caption{Visual comparison of denoising results. (a) Noisy image with $\sigma=30$, (b) result by DDTF method with $p=8$, (c) result by Alg.1 with $p=8$ and $s=20$, (d) result by Alg.1 with $p=8$ and $s=30$, (d) result by DDTF method with $p=16$, (f) result by Alg.1 with $p=16$ and $s=80$.
}
\label{fig4.6}
\end{figure}

Figure \ref{fig4.4} shows the denosing results of different algorithms, and the corresponding learned filters are presented in Figure \ref{fig4.5}. As observed, part of these filters generated by the DDTF method are contaminated by noise terribly due to the influence of high noise. However, these filters are not included in the filters generated by our method. An visual observation of the results for the Lena image can also be obtained in Figure \ref{fig4.6}.

\begin{table} [htbp]
\centering \caption{The comparison of the performance of Algorithm $1$ and the original algorithm \cite{ACHA:ConverageDDTF}}
\scalebox{0.9}{
\begin{tabular}{cc|cccccc}
    \hline
   image & Noise & DDTF\cite{ACHA:DDTF};8 & Alg1(20);8 & Alg1(30);8 & DDTF\cite{ACHA:DDTF};16 & Alg1(80);16 & Alg1(120);16 \\
    \hline
    \multirow{6}{*}{Barbara}
    & 20 & 30.60 & 29.34 & 30.52 & 31.00 & 29.91 & 30.85 \\
    & 30 & 28.45 & 27.84 & 28.57 & 28.94 & 28.58 & 29.03 \\
    & 40 & 26.95 & 26.54 & 27.15 & 27.47 & 27.41 & 27.61 \\
    & 50 & 25.75 & 25.64 & 26.00 & 26.30 & 26.42 & 26.52 \\
    & 60 & 24.70 & 24.84 & 24.97 & 25.39 & 25.61 & 25.47 \\
    & 70 & 23.82 & 24.12 & 24.06 & 24.62 & 24.93 & 24.81 \\
    \hline
    \multicolumn{2}{c|}{Ave. Time (s)} & 4.05 & 1.64 & 2.18 & 54.75 & 18.41 & 26.49 \\
  \hline

    \hline
    \multirow{6}{*}{Hill}
    & 20 & 30.20 & 30.13 & 30.23 & 30.16 & 30.07 & 30.15 \\
    & 30 & 28.54 & 28.65 & 28.66 & 28.52 & 28.56 & 28.60 \\
    & 40 & 27.33 & 27.49 & 27.53 & 27.40 & 27.47 & 27.55 \\
    & 50 & 26.43 & 26.72 & 26.70 & 26.56 & 26.68 & 26.74 \\
    & 60 & 25.68 & 26.05 & 25.99 & 25.82 & 26.09 & 26.09 \\
    & 70 & 25.02 & 25.49 & 25.44 & 25.22 & 25.51 & 25.57 \\
    \hline
    \multicolumn{2}{c|}{Ave. Time (s)} & 4.05 & 1.63 & 2.19 & 55.08 & 18.35 & 26.48 \\
  \hline

    \hline
    \multirow{6}{*}{Lena}
    & 20 & 32.31 & 32.41 & 32.46 & 32.38 & 32.44 & 32.46 \\
    & 30 & 30.31 & 30.58 & 30.59 & 30.47 & 30.63 & 30.64 \\
    & 40 & 28.83 & 29.23 & 29.18 & 29.09 & 29.31 & 29.33 \\
    & 50 & 27.64 & 28.16 & 28.07 & 28.02 & 28.35 & 28.35 \\
    & 60 & 26.67 & 27.22 & 27.14 & 27.10 & 27.47 & 27.45 \\
    & 70 & 25.84 & 26.47 & 26.38 & 26.32 & 26.80 & 26.76 \\
    \hline
    \multicolumn{2}{c|}{Ave. Time (s)} & 4.05 & 1.64 & 2.18 & 55.37 & 18.40 & 26.47 \\
  \hline

    \hline
    \multirow{6}{*}{Man}
    & 20 & 30.02 & 29.81 & 29.98 & 29.79 & 29.62 & 29.76 \\
    & 30 & 28.20 & 28.27 & 28.31 & 27.98 & 27.98 & 28.07 \\
    & 40 & 26.97 & 27.15 & 27.15 & 26.84 & 26.86 & 26.90 \\
    & 50 & 26.05 & 26.28 & 26.27 & 25.96 & 26.10 & 26.16 \\
    & 60 & 25.29 & 25.62 & 25.55 & 25.26 & 25.47 & 25.46 \\
    & 70 & 24.64 & 25.05 & 24.99 & 24.71 & 24.96 & 24.98 \\
    \hline
    \multicolumn{2}{c|}{Ave. Time (s)} & 4.08 & 1.64 & 2.18 & 55.24 & 18.42 & 26.45 \\
  \hline

    \hline
    \multirow{6}{*}{Fingerprint}
    & 20 & 28.36 & 28.42 & 28.44 & 28.36 & 28.37 & 28.41 \\
    & 30 & 26.17 & 26.29 & 26.28 & 26.26 & 26.32 & 26.33 \\
    & 40 & 24.66 & 24.82 & 24.81 & 24.86 & 24.90 & 24.92 \\
    & 50 & 23.47 & 23.65 & 23.64 & 23.82 & 23.88 & 23.95 \\
    & 60 & 22.35 & 22.63 & 22.63 & 22.97 & 23.04 & 23.11 \\
    & 70 & 21.56 & 21.82 & 21.77 & 22.28 & 22.39 & 22.43 \\
    \hline
    \multicolumn{2}{c|}{Ave. Time (s)} & 4.10 & 1.66 & 2.19 & 55.47 & 18.69 & 26.78 \\
  \hline

   \hline
    \multirow{6}{*}{Boat}
    & 20 & 30.35 & 30.19 & 30.40 & 30.24 & 30.13 & 30.25 \\
    & 30 & 28.39 & 28.48 & 28.53 & 28.33 & 28.43 & 28.44 \\
    & 40 & 27.00 & 27.21 & 27.21 & 27.01 & 27.13 & 27.05 \\
    & 50 & 25.95 & 26.24 & 26.17 & 26.00 & 26.16 & 26.20 \\
    & 60 & 25.10 & 25.37 & 25.34 & 25.19 & 25.38 & 25.44 \\
    & 70 & 24.34 & 24.70 & 24.65 & 24.52 & 24.69 & 24.80 \\
    \hline
    \multicolumn{2}{c|}{Ave. Time (s)} & 4.03 & 1.65 & 2.17 & 55.35 & 18.55 & 26.44 \\
  \hline
\end{tabular}}
\label{tab4.2}
\end{table}

In what follows, we compare Algorithm 1 with some of the state-of-the-art denoising methods \cite{TIP:LSSC,CVPR:WNNM} proposed very recently. The codes of these algorithms are all written and implemented in Matlab, and hence the comparison is fair. The numerical experiments in \cite{ACHA:DDTF} demonstrate that the DDTF method runs much faster than the K-SVD method with comparable performance on denoising, thus so is the proposed algorithm. Recently, many patch-based denoising models which utilize the nonlocal similarity of image patches were proved to be superior to the K-SVD method, and hence the proposed algorithm in the denoising performance. However, the corresponding computational amount is much greater. Table \ref{tab4.3} lists the PSNR values and computational time of the three algorithms. Note that ``LASSC" and ``WNNM" represent the models in \cite{TIP:LSSC}\footnote{http://see.xidian.edu.cn/faculty/wsdong/Data/LASSC\_Denoising.rar}
 and \cite{CVPR:WNNM}\footnote{http://www4.comp.polyu.edu.hk/~cslzhang/code/WNNM\_code.zip} respectively. It is observed that the implementation time of the compared methods is too longer than that of Algorithm1, though the denoising performance is better.

\begin{table} [htbp]
\centering \caption{The comparison of the performance of different algorithms: the given values are PSNR (dB)/CPU time(second) }
\scalebox{0.9}{
\begin{tabular}{cccccccc}
  \hline
  \multirow{2}{*}{Image} & \multirow{2}{*}{Noise} & \multicolumn{2}{c}{LASSC}\cite{TIP:LSSC} & \multicolumn{2}{c}{WNNM}\cite{CVPR:WNNM} & \multicolumn{2}{c}{Algorithm 1(30);8}\\
  \cline{3-8}
   &  & PSNR & Time & PSNR & Time & PSNR & Time \\
  \hline\hline
   \multirow{3}{*}{Hill} & $\sigma=20$ & 30.57 & 97.56 & 30.80 & 788.45 & 30.23 & 2.19 \\
  & $\sigma=30$ & 28.91 & 135.11 & 29.20 & 1228.86 & 28.66 & 2.19\\
  & $\sigma=40$ & 27.91 & 271.70 & 28.05 & 1423.83 & 27.53 & 2.19\\
  \hline\hline
   \multirow{3}{*}{Fingerprint} & $\sigma=20$ & 28.97 & 97.56 & 29.05 & 781.6 & 28.44 & 2.22 \\
  & $\sigma=30$ & 26.92 & 136.20 & 27.11 & 1357.58 & 26.28 & 2.18\\
  & $\sigma=40$ & 25.63 & 269.47 & 25.71 & 1415.43 & 24.81 & 2.22\\
  \hline\hline
   \multirow{3}{*}{Man}
  & $\sigma=20$ & 30.53 & 107.10 & 30.73 & 781.62 & 29.98 & 2.22 \\
  & $\sigma=30$ & 28.70 & 152.41 & 29.00 & 1534.87 & 28.31 & 2.19\\
  & $\sigma=40$ & 27.67 & 310.80 & 27.82 & 1467.76 & 27.15 & 2.16\\
  \hline\hline
   \multirow{3}{*}{Boat}
  & $\sigma=20$ & 30.70 & 91.88 & 30.96 & 754.44 & 30.40 & 2.23 \\
  & $\sigma=30$ & 28.83 & 135.16 & 29.16 & 1221.57 & 28.53 & 2.18\\
  & $\sigma=40$ & 27.60 & 263.17 & 27.85 & 1226.50 & 27.21 & 2.19\\
  \hline
\end{tabular}}
\label{tab4.3}
\end{table}

{Finally, in order to further verify the stability of the proposed method, the experiments are conducted on twenty-four standard images randomly selected from a large image dataset which is constructed by the computer vision group of University of Granada. These test images with size of $512\times 512$ (see Figure 6) can be freely downloaded from the network \footnote{http://decsai.ugr.es/cvg/dbimagenes/index.php}. Table \ref{tab4.4} lists the PSNR values of different algorithms. Here ``img. 1-img. 24" denotes the test images in Figure 6 in sequence. Once again we observe that the proposed method can achieve the better PSNRs in most cases. Here we only consider the case of $p=8$. The similar conclusion can be obtained by considering the case of $p=16$.}

\begin{table*} [htbp]
\centering \caption{The comparison of the performance of Algorithm $1$ and the original algorithm \cite{ACHA:ConverageDDTF}}
\scalebox{0.9}{
\begin{tabular}{c|cccc|cccc|cccc}
  \hline
  {Image}  & \multicolumn{4}{c|}{img. 1} & \multicolumn{4}{c|}{img. 2} & \multicolumn{4}{c}{img. 3}\\
  \hline
  {$\sigma$} & 30 & 40 & 50 & 60 & 30 & 40 & 50 & 60 & 30 & 40 & 50 & 60\\
  \hline
  {DDTF\cite{ACHA:DDTF};8}
  & 29.77 & 28.29 & 27.02 & 26.01 & 31.65 & 30.18 & 28.95 & 27.90 & 28.09 & 26.75 & 25.71 & 24.76\\
  \hline
  {Alg1(20);8}
  & 30.00 & 28.62 & 27.42 & 26.39 & 32.07 & 30.70 & 29.59 & 28.72 & 28.19 & 26.94 & 25.95 & 25.02\\
  \hline
  {Alg1(30);8}
  & 29.98 & 28.50 & 27.35 & 26.37 & 32.02 & 30.54 & 29.49 & 28.59 & 28.22 & 26.88 & 25.90 & 24.92\\
  \hline
  \hline
  {Image}  & \multicolumn{4}{c|}{img. 4} & \multicolumn{4}{c|}{img. 5} & \multicolumn{4}{c}{img. 6}\\
  \hline
  {$\sigma$} & 30 & 40 & 50 & 60 & 30 & 40 & 50 & 60 & 30 & 40 & 50 & 60\\
  \hline
  {DDTF\cite{ACHA:DDTF};8}
  & 25.10 & 23.67 & 22.62 & 21.78 & 27.21 & 25.99 & 24.95 & 24.01 & 26.20 & 24.99 & 24.14 & 23.46\\
  \hline
  {Alg1(20);8}
  & 24.84 & 23.58 & 22.63 & 21.84 & 27.19 & 26.06 & 25.09 & 24.18 & 25.42 & 24.63 & 23.99 & 23.44\\
  \hline
  {Alg1(30);8}
  & 25.08 & 23.70 & 22.69 & 21.86 & 27.25 & 26.08 & 25.08 & 24.17 & 25.88 & 24.87 & 24.13 & 23.50\\
  \hline
  \hline
  {Image}  & \multicolumn{4}{c|}{img. 7} & \multicolumn{4}{c|}{img. 8} & \multicolumn{4}{c}{img. 9}\\
  \hline
  {$\sigma$} & 30 & 40 & 50 & 60 & 30 & 40 & 50 & 60 & 30 & 40 & 50 & 60\\
  \hline
  {DDTF\cite{ACHA:DDTF};8}
  & 29.33 & 28.13 & 27.11 & 26.22 & 31.07 & 29.56 & 28.54 & 27.55 & 30.89 & 29.49 & 28.29 & 27.17\\
  \hline
  {Alg1(20);8}
  & 29.50 & 28.41 & 27.44 & 26.64 & 31.42 & 29.94 & 29.05 & 28.21 & 31.12 & 29.88 & 28.78 & 27.78\\
  \hline
  {Alg1(30);8}
  & 29.48 & 28.36 & 27.42 & 26.59 & 31.38 & 29.88 & 28.99 & 28.09 & 31.13 & 29.81 & 28.75 & 27.70\\
  \hline
  \hline
  {Image}  & \multicolumn{4}{c|}{img. 10} & \multicolumn{4}{c|}{img. 11} & \multicolumn{4}{c}{img. 12}\\
  \hline
  {$\sigma$} & 30 & 40 & 50 & 60 & 30 & 40 & 50 & 60 & 30 & 40 & 50 & 60\\
  \hline
  {DDTF\cite{ACHA:DDTF};8}
  & 26.20 & 25.03 & 24.13 & 23.48 & 26.31 & 25.20 & 24.32 & 23.59 & 28.49 & 27.27 & 26.20 & 25.29\\
  \hline
  {Alg1(20);8}
  & 25.95 & 24.95 & 24.17 & 23.56 & 26.27 & 25.23 & 24.48 & 23.76 & 28.54 & 27.45 & 26.43 & 25.62\\
  \hline
  {Alg1(30);8}
  & 26.15 & 25.03 & 24.22 & 23.59 & 26.32 & 25.24 & 24.46 & 23.74 & 28.58 & 27.44 & 26.45 & 25.56\\
  \hline
  \hline
  {Image}  & \multicolumn{4}{c|}{img. 13} & \multicolumn{4}{c|}{img. 14} & \multicolumn{4}{c}{img. 15}\\
  \hline
  {$\sigma$} & 30 & 40 & 50 & 60 & 30 & 40 & 50 & 60 & 30 & 40 & 50 & 60\\
  \hline
  {DDTF\cite{ACHA:DDTF};8}
  & 30.17 & 29.11 & 28.11 & 27.28 & 28.09 & 26.64 & 25.59 & 24.76 & 27.84 & 26.45 & 25.34 & 24.45\\
  \hline
  {Alg1(20);8}
  & 30.28 & 29.38 & 28.56 & 27.91 & 28.14 & 26.80 & 25.78 & 25.06 & 27.40 & 26.32 & 25.39 & 24.59\\
  \hline
  {Alg1(30);8}
  & 30.32 & 29.37 & 28.52 & 27.83 & 28.19 & 26.80 & 25.77 & 24.98 & 27.75 & 26.53 & 25.48 & 24.64\\
  \hline
  \hline
  {Image}  & \multicolumn{4}{c|}{img. 16} & \multicolumn{4}{c|}{img. 17} & \multicolumn{4}{c}{img. 18}\\
  \hline
  {$\sigma$} & 30 & 40 & 50 & 60 & 30 & 40 & 50 & 60 & 30 & 40 & 50 & 60\\
  \hline
  {DDTF\cite{ACHA:DDTF};8}
  & 27.84 & 26.59 & 25.47 & 24.62 & 25.19 & 24.12 & 23.35 & 22.75 & 24.23 & 22.92 & 22.06 & 21.37\\
  \hline
  {Alg1(20);8}
  & 27.91 & 26.76 & 25.74 & 24.86 & 24.99 & 24.07 & 23.37 & 22.78 & 23.61 & 22.66 & 21.95 & 21.34\\
  \hline
  {Alg1(30);8}
  & 27.95 & 26.70 & 25.69 & 24.84 & 25.12 & 24.13 & 23.36 & 22.82 & 24.01 & 22.86 & 22.06 & 21.42\\
  \hline
  \hline
  {Image}  & \multicolumn{4}{c|}{img. 19} & \multicolumn{4}{c|}{img. 20} & \multicolumn{4}{c}{img. 21}\\
  \hline
  {$\sigma$} & 30 & 40 & 50 & 60 & 30 & 40 & 50 & 60 & 30 & 40 & 50 & 60\\
  \hline
  {DDTF\cite{ACHA:DDTF};8}
  & 23.24 & 22.09 & 25.47 & 21.30 & 30.02 & 28.92 & 28.02 & 27.19 & 26.92 & 26.04 & 25.37 & 24.85\\
  \hline
  {Alg1(20);8}
  & 22.50 & 21.72 & 25.74 & 21.11 & 30.18 & 29.25 & 28.48 & 27.80 & 26.91 & 26.07 & 25.50 & 25.07\\
  \hline
  {Alg1(30);8}
  & 22.84 & 21.89 & 25.69 & 21.21 & 30.19 & 29.23 & 28.42 & 27.71 & 26.97 & 26.10 & 25.52 & 25.07\\
  \hline
  \hline
  {Image}  & \multicolumn{4}{c|}{img. 22} & \multicolumn{4}{c|}{img. 23} & \multicolumn{4}{c}{img. 24}\\
  \hline
  {$\sigma$} & 30 & 40 & 50 & 60 & 30 & 40 & 50 & 60 & 30 & 40 & 50 & 60\\
  \hline
  {DDTF\cite{ACHA:DDTF};8}
  & 29.38 & 27.99 & 26.80 & 25.93 & 26.53 & 25.23 & 24.31 & 23.68 & 26.19 & 24.98 & 24.03 & 23.27\\
  \hline
  {Alg1(20);8}
  & 29.22 & 28.07 & 27.02 & 26.23 & 26.28 & 25.21 & 24.38 & 23.84 & 26.12 & 25.03 & 24.15 & 23.44\\
  \hline
  {Alg1(30);8}
  & 29.35 & 28.14 & 27.02 & 26.24 & 26.50 & 25.27 & 24.42 & 23.76 & 26.22 & 25.07 & 24.15 & 23.39\\
  \hline

\end{tabular}}
\label{tab4.4}
\end{table*}

\section{Conclusion}\label{sec5}

In this paper, inspired by the fact that part of the data-driven tight frame filters learned by the recently proposed method \cite{ACHA:DDTF} are influenced by the noise severely, we propose an improved data-driven filters learning method. In our method, we divide the matrix used for generating the tight frame filters into two part, corresponding to the so called signal subspace and noise subspace. Then only the bases spanning the signal subspace are used for constructing the learned filters. This means that the filters contaminated by the noise severely are excluded for the sparsity representation of images. Numerical experiments demonstrate that our method overall outperforms the original data-driven tight frame construction scheme.

\section*{acknowledgement}
The work was supported in part by the National Natural Science Foundation of China under Grant 61271014 and 61401473.

\end{document}